\makeatletter\def\graphicscache@inhibit{true}\makeatother
\documentclass[twocolumn, 9pt, a4paper]{scrartcl}

\usepackage{amssymb, amsthm, amsmath}
\usepackage{caption}
\usepackage{subcaption}
\usepackage{algorithm}
\usepackage{algpseudocode}
\usepackage{bm}
\usepackage{stfloats}
\usepackage{placeins}
\usepackage{tablefootnote}
\usepackage{graphicx}
\usepackage{multirow}
\usepackage{graphicscache}
\usepackage{lineno,hyperref}
\usepackage{siunitx}
\usepackage{tikz}
\usepackage{fancyhdr}
\usepackage{setspace}
\floatstyle{plaintop}
\restylefloat{table}

\usepackage[letterspace=-15]{microtype}

\newtheorem{lemma}{Lemma}

\newcommand{\ftof}{\texttt{f2f}}
\newcommand{\ftokf}{\texttt{f2kf}}
\newcommand{\ftor}{\texttt{f2r}}

\newcommand{\figref}[1]{\mbox{Figure}~\ref{#1}}
\newcommand{\tabref}[1]{\mbox{Table}~\ref{#1}}
\newcommand{\etal}{et al.}
\newcommand{\secref}[1]{\mbox{Section}~\ref{#1}}
\usepackage{xpatch}
\xpretocmd{\eqref}{Eq.~}{}{}
\makeatletter
\newcommand\footnoteref[1]{\protected@xdef\@thefnmark{\ref{#1}}\@footnotemark}
\makeatother

\DeclareMathOperator{\SE3}{\mathbf{SE}(3)}
\DeclareMathOperator{\Sim3}{\mathbf{Sim}(3)}
\DeclareMathOperator{\se3}{\mathfrak{se}(3)}
\DeclareMathOperator{\sim3}{\mathfrak{sim}(3)}

\title{Rendering the Directional TSDF for Tracking and Multi-Sensor Registration with Point-To-Plane Scale ICP}
\author{Malte Splietker and Sven Behnke}
\date{\vspace{-5ex}}

\begin{document}

\maketitle

\thispagestyle{empty}

\begin{tikzpicture}[remember picture,overlay]
  \node[anchor=north,align=center,font=\sffamily,yshift=-0.2cm] at (current page.north) {%
  Published in Robotics and Autonomous Systems, Elsevier, 2023.\\\url{https://doi.org/10.1016/j.robot.2022.104337}
  };
\end{tikzpicture}

\begin{abstract}
  \normalfont%
  \lsstyle%
  \begin{spacing}{0.95} %
  \bfseries\textit{Abstract}---\,%
Dense real-time tracking and mapping from \mbox{RGB-D} images is an important tool for many robotic applications, such as navigation and manipulation.
The recently presented Directional Truncated Signed Distance Function (DTSDF) is an augmentation of the regular TSDF that shows potential for more coherent maps and improved tracking performance. In this work, we present methods for rendering depth- and color images from the DTSDF, making it a true drop-in replacement for the regular TSDF in established trackers. We evaluate the algorithm on well-established datasets and observe that our method improves tracking performance and increases re-usability of mapped scenes. Furthermore, we add color integration which notably improves color-correctness at adjacent surfaces.
Our novel formulation of combined ICP with frame-to-keyframe photometric error minimization further improves tracking results.
Lastly, we introduce $\Sim3$ point-to-plane ICP for refining pose priors in a multi-sensor scenario with different scale factors.
  \end{spacing}
  \normalfont\normalsize
\end{abstract}

\section{Intoduction}
\label{sec:introduction}
Since its first appearance in KinectFusion~\cite{Newcombe2011}, GPU accelerated TSDF algorithms have become a de-facto standard in scene reconstruction from depth images, leveraging inexpensive sensors and massive parallel processing on GPUs for good real-time performance.
By modeling the closest distance to the next surface with a signed distance function (SDF), geometry can be reconstructed by finding the zero-transition from positive (i.e., in front of the surface) to negative (i.e., behind the surface) values. In practice, this function is obtained by fusing measurements into a regular grid, the so called voxels, and interpolating between them.
The necessity to store both front- and backside of the surface implies, however, that there is a minimum thickness of objects that can be represented. Especially with thin objects, integration of new measurements might interfere with and contradict old data belonging to a different surface, leading to a corrupted model. We have explored this issue and introduced the concept of the Directional Truncated Signed Distance Function (DTSDF) in our previous work~\cite{Splietker2019}.
The DTSDF\footnote{Code available at \url{https://github.com/AIS-Bonn/DirectionalTSDF}} uses six TSDF volumes, one for each positive and negative coordinate axis, to store surface sections with different orientations. We proposed a method for fusing depth images into this data structure and for extracting meshes with a modified marching cubes algorithm. The latter is, however, not applicable for real-time tracking applications.
\begin{figure}[!h]
  \centering
  \includegraphics[width=\linewidth]{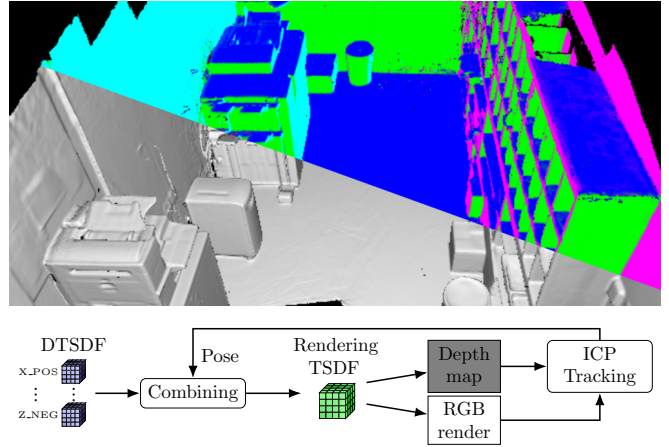}
  \caption{The top part shows a cut view of the reconstructed Stanford copyroom scene. The lower left half is the rendered depth; upper right shows the directions involved in rendering in different colors, mixed whenever multiple directions contributed. The bottom part shows the pipeline for rendering and tracking with the DTSDF.}
  \label{fig:iconic}
\end{figure}

Instead, in this work we propose methods for rendering virtual camera views, which allows to use the standard ICP algorithm for real-time sensor motion tracking. Moreover, we introduce color integration into the DTSDF, which helps preserving color details of adjacent object surfaces, especially along sharp edges. With these additions, the DTSDF becomes a true replacement for the regular TSDF with only minor modifications to the fusion and rendering algorithms. We showcase and evaluate our method on well-known datasets and deduce, that the DTSDF has advantages in tracking certain types of sequences and is better at preserving the overall map for later reuse.

Throughout the related work the ICP algorithm has been implemented with some variations, some of which we discuss in \secref{sec:icp_tracking}. Especially, we investigate the use of photometric ICP with the TSDF and found, that there are different variations of frame-to-frame and frame-to-model photometric ICP, but no frame-to-keyframe. We re-formulate the error function accordingly and conduct a case study which indicates, that frame-to-keyframe is favorable in most sequences.

As part of the PhenoRob Cluster of Excellence~\footnote{Cluster of Excellence PhenoRob -- Robotics and Phenotyping for Sustainable Crop Production \url{https://www.phenorob.de}} we are involved in a project to create 3D reconstructions of crops captured by a robot on the field. One of the challenges is to align the multiple point clouds from high-accuracy laser depth- and RGB stereo sensors. Even given an offline pose prior of the sensors, each scan requires pose refinement, due to the limited rigidity of the robot's chassis.
The computed depth from the stereo-pairs is also affected by geometric changes, so in \secref{sec:sim3_icp} we showcase our method by extending the point-to-plane $\SE3$ ICP algorithm to the $\Sim3$ Group, which jointly optimizes translation, rotation and scale.

\section{Related Work}
\label{sec:related_work}
In 3D reconstruction and SLAM feature-based, sparse, and dense methods are distinguished. Our work belongs to the category of dense methods that describe closed surfaces, separate objects, and even free space~\cite{Oleynikova2017}.

Research focus in this field has shifted in recent years towards learned representations. Occupancy networks~\cite{Mescheder2019a} learn a binary descriptor describing the occupancy of space, i.e., whether a point lies inside an object or not. In DeepSDF by Park \etal~\cite{Park2019}, a representation is learned, which like our work allows querying the signed distance to the closest object for arbitrary points in space. Neural Radiance Fields (NeRF)~\cite{Mildenhall2020, Azinovic2021} use deep networks to regress density and color. While these approaches show impressive results and use less memory to store the model or even enable scene completion, they have some shortcomings. The limited model size results in a lack of detail for large scenes. Also training and inference times, though improving lately, are still not applicable for real-time applications. Hence, the classic TSDF fusion algorithms are still state-of-the-art in live mapping scenarios.

Since its first occurrence, the TSDF fusion algorithm has seen widespread use cases and is mostly used in its original form without changes to the representation. There have been attempts to augment it, though. Dong \etal~\cite{Dong2018a} created a hybrid data structure, combining the TSDF with probabilistic surfels. Multiple overlapping TSDF sub-volumes are used in pose graphs for large-scale SLAM, enabling re-aligning parts of the map on loop closure detection for consistency~\cite{Henry2013, Whelan2015, Millane2018}. This approach is similar to the DTSDF in the way it maintains several overlapping representations, but does not fix the TSDF's inability to represent thin objects observed from opposite sides within the same volume. While this may not be an issue for some applications, object scans and walk-around type scenes profit from this capability.

Zhang \etal~\cite{Zhang2021} give an overview of current RGB-D SLAM algorithms. %
The typical method for localizing the sensor pose in the TSDF is frame-to-model geometric ICP~\cite{Newcombe2011}, where the current depth image is registered against a point cloud rendered from the TSDF at the previous position. It uses the point-to-plane metric and Gauss Newton for minimizing the sum of squared errors. There are modified versions, such as the extended ICP tracker~\cite{Prisacariu2017}, which uses the Huber-norm and has advanced outlier detection. Nguyen \etal~\cite{Nguyen2012} model the depth dependent noise of the RGB-D sensor in a weighted ICP scheme. Xia \etal~\cite{Xia2018} use a simplified weight based on the inverse quadratic depth. But the regular ICP is still most commonly used.

Photometric ICP instead uses a registration loss that is based on the per-pixel photometric error of RGB images~\cite{Steinbruecker2011}. This is helpful for preventing drift in geometrically ambiguous scenes, e.g., textured planar surfaces. To take advantage of both, color and depth, combined ICP optimizes both error functions simultaneously~\cite{Kerl2013, Henry2013, Whelan2015, Dai2017}.

Common for most TSDF ICP implementations is, that they register the current input depth- and color images against point clouds and color images rendered from the TSDF (frame-to-render). Some combined ICP variants use frame-to-render for geometric- and frame-to-frame for photometric registration~\cite{Whelan2015, Prisacariu2017}. Moreover, there is the direct volume matching line of algorithms that directly performs registration within the SDF. Point-to-SDF~\cite{Bylow2016, Canelhas2013, Palazzolo2019} and SDF-to-SDF~\cite{Slavcheva2018} approaches can be distinguished. Millane \etal~\cite{Millane2020} recently proposed a method for extracting and matching local features directly on the SDF.
Model-less RGB-D SLAM systems like DVO~\cite{Kerl2013} or ORB-SLAM~\cite{Mur-Artal2015} often use keyframes combined with a pose graph. Even without graph optimization, the use of keyframes has an advantage over frame-to-frame tracking which tends to accumulate drift faster. To our knowledge, this has not been used in combination with the TSDF yet. \secref{sec:photometric_icp} discusses and compares the different option choices.

Further hybrid approaches utilize other tracking sources. BundleFusion~\cite{Dai2017} is an advanced method that combines ICP error minimization and visual SIFT features in a global bundle adjustment, and then de- and reintegrates parts of scene to keep the overall representation consistent.

The goal of this work is to make DTSDF usable as replacement or supplement for the regular TSDF. To be able to profit from established tracking methods without further modifications we
\begin{itemize}
  \item introduce color fusion into the DTSDF,
  \item present an efficient method for generating rendered views of the DTSDF,
  \item use these rendered views to track sensor motion with the combined ICP algorithm,
  \item derive a formulation for combined ICP with keyframes for the photometric error,
  \item Implement $\Sim3$ point-to-plane ICP for jointly optimizing pose and scales in a multi-sensor setup.
\end{itemize}

\section{Fusion and Weights}
\label{sec:fusion_and_weights}
Formally, the signed distance function $\Phi:\mathbb{R}^3 \longrightarrow (d, w_d, \mathbf{c}, w_c)$ maps an arbitrary point in space to a tuple comprising signed distance to the closest surface $d$, distance weight $w_d$, RGB color $\mathbf{c}$ and color weight $w_c$, where the weights represent the confidence of the integrated information. The surfaces of the environment are determined by finding zero-transition of $\Phi$, i.e., finding the subset in $\mathbb{R}^3$ where the signed distance turns from positive (in front of surface) to negative (behind surface) values. As reconstruction only requires information close to the actual surface, the TSDF only maps points within a truncation band $\tau$ around the actual surface.
In practice, the TSDF is stored as a evenly-spaced grid of voxels and the signed distance, color, and weights in $\Phi$ for an arbitrary point in $\mathbb{R}^3$ are estimated by linear interpolation between tuples stored in the neighboring voxels.

The directional TSDF~\cite{Splietker2019} $\Phi_{\mathrm{dir}}(\mathbf{p}) = (\Phi^D(\mathbf{p}))_{D\in\textrm{Directions}}$ extends this representation by mapping a point to multiple signed distance functions -- one for each direction $\{X^+, X^-, Y^+, Y^-, Z^+, Z^-\}$ -- corresponding to the positive and negative coordinate axes $\mathbf{v} = \{(1, 0, 0)^\intercal, \cdots, (0, 0, -1)^\intercal \}$. Observed depth points are assigned to those directions $D$ that fulfill
\begin{equation}
  \arccos \langle \mathbf{n}, \mathbf{v}^D \rangle < \theta
  \label{eq:angular_threshold}
\end{equation}
for depth normal $\mathbf{n}$ and angular threshold $\theta \in (\pi / 4, \pi /2]$, i.e., the angular difference between surface normal and direction vector is smaller than threshold $\theta$. The range for $\theta$ is chosen, such that an overlap between neighboring directions is guaranteed and every surface point matches at least one and at most three directions.

\begin{figure}
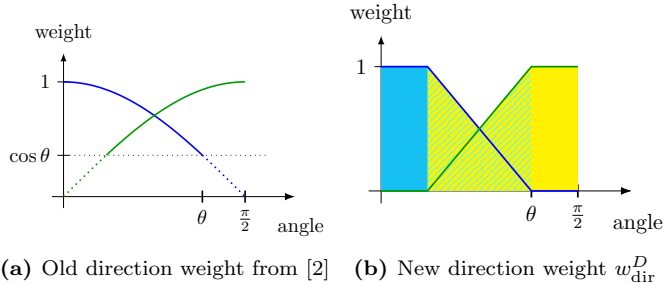

  \centering
  \setlength{\belowcaptionskip}{0pt}
  \begin{subfigure}[b]{0.485\linewidth}
    \includegraphics[width=\linewidth]{images/direction_weight_old.pdf}
    \caption{Old direction weight from~\cite{Splietker2019}}
    \label{fig:direction_weight_old}
  \end{subfigure}
  \hfill
  \begin{subfigure}[b]{0.485\linewidth}
    \includegraphics[width=\linewidth]{images/direction_weight.pdf}
    \caption{New direction weight $w_{\textrm{dir}}^D$}
    \label{fig:direction_weight_new}
  \end{subfigure}
  \setlength{\belowcaptionskip}{-5pt}
  \caption{Direction weight for angle $\arccos \langle\mathbf{n}, \mathbf{v}^D\rangle$ between surface normal $\mathbf{n}$ and two neighboring direction vectors (e.g., $\mathbf{v}^{X^+}$ and $\mathbf{v}^{Y^+}$) indicated by the blue and green lines, respectively. The angular threshold $\theta$ determines the width of exclusive (solid) and overlapping (hatched) areas of the two directions.}
  \label{fig:direction_weight}
\end{figure}

Fusion is the process of integrating new observations into the voxels as weighted cumulative moving average, where weights denote the certainty of the added information. For every voxel and associated depth point, a fusion weight is calculated. While there is no definite weighting scheme, most implementations use a combination of factors to compensate for measurement noise and for uncertainty by down-weighting voxels behind the surface \cite{Newcombe2011, Bylow2013, Nguyen2012}.
In addition to these factors, the fusion weight in~\cite{Splietker2019} includes a direction factor $\hat{w}_{\textrm{dir}}^D = \left\langle \mathbf{n}, \mathbf{v}^D \right\rangle$ to blend surfaces which are represented by multiple directions (c.f. \eqref{eq:angular_threshold}) over the whole span of $[0, \pi / 2]$. Then, all $\hat{w}_{\textrm{dir}}^D$ are explicitly compared to threshold $\cos(\theta)$ and fused, if smaller.
Instead, we propose the membership function
\begin{align}
  w_{\textrm{dir}}^D(\mathbf{n}) =
  \min\left(\max\left(
  \frac{1 - \arccos \langle \mathbf{n}, \mathbf{v}^D \rangle}{2 \theta - \frac{\pi}{4}}, 0
  \right), 1\right)
  \label{eq:direction_weight}
\end{align}
  which has multiple advantages, as illustrated in~\figref{fig:direction_weight}. Firstly, for angles larger than $\theta$ the weight becomes zero, so explicit thresholding becomes superfluous. In the exclusive area, where the point belongs to exactly one direction, the weight is one. The blending area, where a point belongs to multiple directions, now blends linearly on the full $[0, 1]$ range, whereas the old function had an effective range of $[\cos \theta, 1]$.
Just like in the previous work~\cite{Splietker2019}, the combined depth fusion weight for fusing a point into direction $D$ is
\begin{align}
  w_d = w_{\mathrm{depth}} \cdot w_{\mathrm{angle}} \cdot w_{\textrm{dir}}^D.
  \label{eq:combined_fusion_weight}
\end{align}
The weight function parameters were omitted for improved readability.

Color is fused analogous to distances, but with a different weight. Again, there are different variants throughout literature. Dryanovski~\etal~\cite{Dryanovski2017} use the same constant weight for depth and color to save computation time.
Whelan \etal~\cite{Whelan2015} use angle between depth normal and view ray to downweight steep observation angles, which Bylow~\etal~\cite{Bylow2013} use in combination with the depth weight. This factor is also included in our depth weight.
We argue that using the depth fusion weight for colors is important, as the uncertainty is reflected in the choice of voxels associated with pixels. Let $\mathbf{x} \in \mathbb{R}^3$ be the voxel location and $\mathbf{p} \in \mathbb{R}^3$ the depth point with associated color that is fused into the voxel. Then our color fusion weight is
\begin{align}
  w_c = w_d \left(1 - \min\left(1, \frac{||\mathbf{p} - \mathbf{x}||}{\tau}\right) \right),
  \label{eq:color_weight}
\end{align}
where $w_d$ is the depth fusion weight. The factor in parenthesis reduces the confidence for voxels further away from the surface, as multiple colors from various observations may blend together here. For depth fusion we use the point-to-plane metric, which mitigates this issue, but there is no equivalent for colors, as their information is only accurate right at the surface.

Free space, that is space between camera and observed surface, is not explicitly mapped to save memory. Nonetheless, due to noise, sensor error, or dynamic objects it can happen that spurious measurements are mapped in space, that is unoccupied in reality, and it is important to remove these artifacts.
When the computed distance~\eqref{eq:point-to-plane} is larger than the truncation range $\tau$, the voxel is located in free space and updated with a SDF value of 1 and a constant weight. No directional weight is used in this case, as the goal is to carve everything in free space. Special care has to be taken at depth discontinuities: carving can corrupt voxels of edges, because aliasing and small tracking inaccuracies associate the voxel with a more distant surface. Therefore, carving is only applied if there is no depth difference of more than $\tau$ in a radius of two pixels to the associated depth pixel.
To free up memory, voxels that are erroneously allocated but become free space by repeated carving can be recycled in an asynchronous process, as has been done in~\cite{Dong2018}.

For depth point $\mathbf{p}$ with normal $\mathbf{n}$ and truncation range $\tau$, the signed distance to voxel position $\mathbf{x}$ is computed with the point-to-plane metric
\begin{align}
  d = \frac{1}{\tau} \langle \mathbf{p} - \mathbf{x}, \mathbf{n} \rangle.
  \label{eq:point-to-plane}
\end{align}
For convenience, the distance is normalized and clamped to $[-1, 1]$. The point-to-plane metric helps keeping the SDF consistent with varying observation angles as opposed to the point-to-point metric~\cite{Bylow2013}.

While in our previous work~\cite{Splietker2019} we explored ray casting similarly to Klingensmith \etal~\cite{Klingensmith2015} for fusing individual depth pixels along the view- or normal direction, this method often shows bad results with noisy real-world data. For tracking applications, voxel projection, like in the original KinectFusion, has proven more robust. During voxel projection fusion, every allocated voxel within the current view frustum is projected into the current camera frame, associated with a depth (and color) pixel, and updated with the respective values and weights.

\section{DTSDF Raycast Rendering}
\label{sec:dtsdf_rendering}
Rendering real-time views of the model from arbitrary positions is useful for visualization and also tracking. 
Instead of developing specialized tracking methods for the DTSDF, our approach is to render a map of depth points and use known and tested ICP-based algorithms~\cite{Newcombe2011, Prisacariu2017, Kerl2013, Henry2013} to register input depth images.
\begin{figure*}
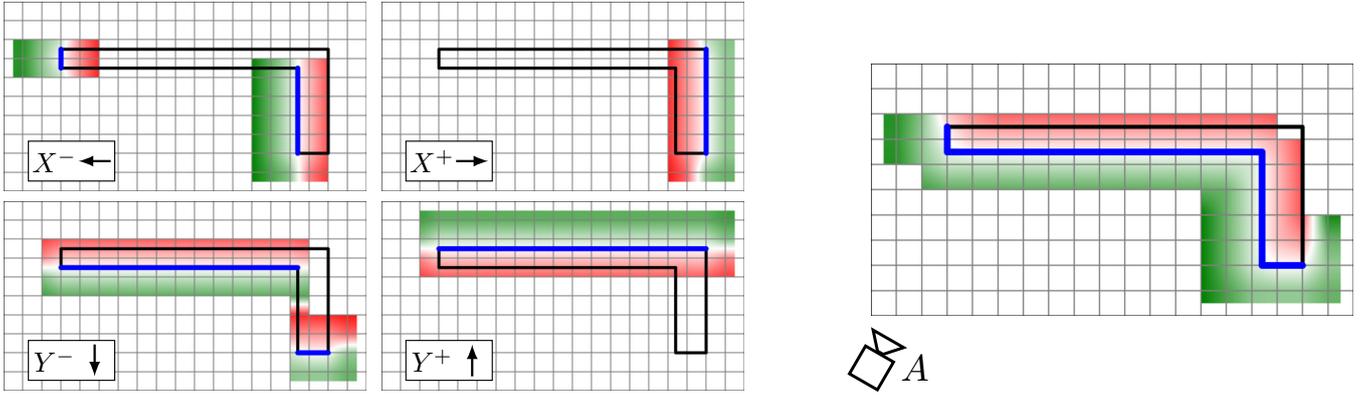

  \setlength{\belowcaptionskip}{0pt}
  \subcaptionbox{TSDFs of directions $X^+, X^-, Y^+, Y^-$ for the scene. Blue indicates surfaces, that fulfill $w_\textrm{dir}^{D}(\mathbf{n}) > 0$ (c.f. \eqref{eq:direction_weight}), i.e., surfaces that are represented by direction $D$.\label{fig:positional_unambiguity_dir}}
    [.55\linewidth]
  {
    \centering
    \includegraphics[width=\linewidth]{images/combination_new_directions.pdf}
  }
  \hfill
  \subcaptionbox{combined TSDF from view point $A$. The blue line indicates the surfaces that are visible from $A$. \label{fig:positional_unambiguity_combined}}[.42\linewidth]
  {
    \centering
    \includegraphics[width=\linewidth]{images/combination_new_combined.pdf}
  }
  \setlength{\belowcaptionskip}{-5pt}
  \caption{2D example of computing the combined TSDF. The black outline represents the mapped object, green/red gradients correspond to positive/negative SDF values and the grid denotes the voxel grid.}
  \label{fig:positional_unambiguity}
\end{figure*}

The rendering process involves casting a ray per pixel of the virtual depth camera and extracting the iso-surface, i.e., the first transition from positive to negative SDF values. This involves probing the TSDF along the ray at regular intervals, until the distance turns negative and then multiple small steps, determined by the interpolated SDF value, to minimize the absolute distance value.
Similar to the meshing presented in~\cite{Splietker2019} the question is: how to combine up to six SDF values from partially overlapping directions?

By its mathematical definition, the signed distance function can represent any given object. In other words, the six directions could, given a fine enough resolution, be combined into a single, conflict-free TSDF. But in practice, the combination is not straight forward: overlapping free and occupied space from different volumes has to be combined in accordance with the orientation of mapped surfaces, while considering corner cases, real-world noise and imperfections.
Also, the practical use is limited. Ray-cast rendering relies on the width of the truncation range for finding zero-transitions, which for thin objects can be easily missed.
Instead, we made an important observation:
\begin{lemma}
  For a DTSDF and a fixed camera position, a combined conflict-free regular TSDF can be computed.
  \label{lem:positional_unambiguity}
\end{lemma}
The basic intuition behind this lemma is, that surfaces the camera perceives from the backside are not relevant from a given position. \figref{fig:positional_unambiguity} visualizes this idea as a 2D example, omitting the z-axis: a thin, L-shaped object is represented by TSDFs for the directions $X^-, X^+, Y^-$ and $Y^+$ as depicted in \figref{fig:positional_unambiguity_dir}. 
For the given resolution, a complete representation by a regular TSDF is not possible, as the negative distance for the inside of the object cannot be stored. However, given a camera pose $A$, a conflict-free combination is possible, as depicted in \figref{fig:positional_unambiguity_combined}. Only the blue surfaces are visible and negative SDF values from backside surfaces are omitted. Positive values are still important, as they prevent rendering surfaces in free space, like the overhanging zero-transitions of individual directions.
In comparison, even doubling the resolution barely enables visualization, as the inside of the object is very thin. This is especially problematic, when the depth noise and truncation range exceed the thickness.

The challenge is, for each point in $p \in \mathbb{R}^3$ to decide whether it is free space (SDF $>$ 0) or occupied (SDF $\leq$ 0) and which directions to combine. We developed the following algorithm:
\begin{algorithm}[!h]
\caption{Compute Combined TSDF.}\label{alg:cap}
\begin{algorithmic}
  \Procedure{CombinedTSDF}{$\mathbf{p}, \mathbf{c}$}
  \State $\mathbf{r} \gets \frac{\mathbf{p} - \mathbf{c}} {||\mathbf{p} - \mathbf{c}||}$ \Comment{view ray}
  \State $\mathrm{freespace} \gets 0$
  \If{$\exists D: \Phi^D(\mathbf{p}) > 0$}
  \State $\mathbf{q} \gets$ first zero-transition from $\mathbf{p}$ in direction $-\mathbf{r}$ in $\Phi^D$
  \If{$\mathbf{q} = \emptyset$}
      \State $\mathrm{freespace} \gets 1$
      \ElsIf{$\not \exists \widetilde{D}: \Phi^{\widetilde{D}}(\mathbf{q}) < 0$ \textbf{and} $\langle \mathbf{n}^{\widetilde{D}}(\mathbf{q}), -\mathbf{r} \rangle > 0$}
      \State $\mathrm{freespace} \gets 1$
    \EndIf
  \EndIf
  \If{$\mathrm{freespace} = 1$}
    \Return weighted sum of free space\\\hspace{4.8cm}directions
  \Else \  %
    \Return weighted sum of occupied space directions
  \EndIf
  \EndProcedure
\end{algorithmic}
\end{algorithm}

\begin{figure}[t]
  \centering
  \includegraphics[scale=1.2]{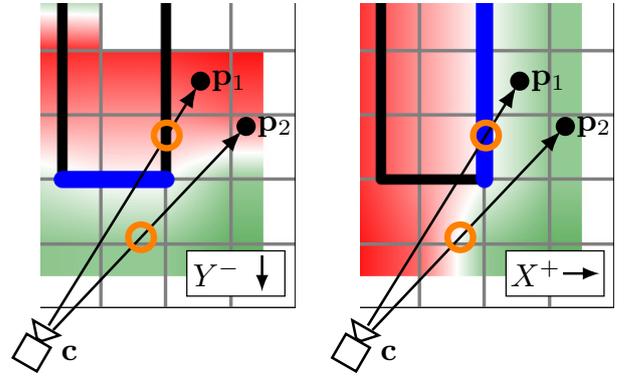}
  \caption{Visualization of view-point dependent free-space detection. From the camera's perspective point $\mathbf{p}_1$ is hidden behind the surface. The zero-transition (circles) in $X^+$ found by ray-casting from $\mathbf{p}_1$ towards $\mathbf{c}$ lies inside the occupied space of $Y^-$, so there is no free space. The first zero-transition from $\mathbf{p}_2$, on the other hand, lies within the free space of $Y^-$, so it does not conflict and $\mathbf{p}_2$ is marked as free space. The same holds, if there is no zero transition for $\mathbf{p}_2$ (i.e., $q = \emptyset$). }
  \label{fig:free_space_detection}
\end{figure}

The input values are the point to look up $\mathbf{p}$ and camera position $\mathbf{c}$. If there is a direction $D$, for which $\mathbf{p}$ lies in free space it has to be checked, whether it conflicts with other directions under consideration of the cameras position: if this free space in $D$ lies within the occupied space of a visible surface from another direction $\widetilde{D}$ and, moreover, the surface point $q$ on the ray between $\mathbf{p}$ and $\mathbf{c}$ in $\widetilde{D}$ also lies within the occupied space, $\mathbf{p}$ cannot be free space. \figref{fig:free_space_detection} illustrates explain this rule with a positive and negative example, each. After determining whether $\mathbf{p}$ is free space or occupied space, all congruent directions are combined as weighted sum using the fused voxel weights.

There are some additional considerations for implementing the algorithm, including
\begin{itemize}
  \item ignoring information from points with invalid gradients w.r.t. the direction vector, as this is not supposed to be represented by that direction.
  \item If point $\mathbf{p}$ is close to a surface and one direction indicates free space, one occupied space, the algorithm would only consider free space, instead of a weighted combination, even when both directions map the same surface.
  \item In reality there are many TSDF entries with low weight, which can erroneously induce free space.
\end{itemize}

While in theory and on synthetic data fused from ground truth poses the algorithm performs well, real-world usage is very limited. Due to sensor noise and slight tracking errors, fused data often leads false free space identification, resulting in dents and holes in the rendered surfaces --- especially near corners. Instead, we propose a simple weighting scheme as approximation, which performs better in practice:

For a point $\mathbf{x} \in \mathbb{R}^3$ and direction $D$, let $\mathbf{n}^D$ be the normalized SDF gradient $(\partial \Phi^D / \partial x,\partial \Phi^D / \partial y,\partial \Phi^D / \partial z)^\intercal$ at $\mathbf{x}$, $w_d^D$ the stored distance weight and $\mathbf{r}$ the normalized view ray from camera center to $\mathbf{x}$. Then the combination weight for direction $D$ is defined as
\begin{align}
  w_{\mathrm{comb}}^D = 
  w_{\textrm{dir}}^D(\mathbf{n}^D)
  \cdot
  \langle \mathbf{n}^D, -\mathbf{r} \rangle
  \cdot
  w_d^D.
  \label{eq:combine_weight}
\end{align}
The first factor in \eqref{eq:combine_weight} ensures that only gradients that actually comply with the direction they are stored in are used, with according weights to blend between directions. The second factor ensures that only directions with eligible surfaces are used, which is the main reason for using the DTSDF. The approximation is not perfect and certain constellations work only under the premise, that all direction's SDF weights are similar. On the other hand, it has shown to be more robust in practice than the algorithm proposed above.

These per-direction weights can be used to directly look up the combined SDF value at any point in space as weighted sum, but ray-casting becomes very slow, because many TSDF lookups and memory reads have to be performed, especially for the gradient computation. The massive parallel computation also results in many cache misses, so the algorithm becomes memory bound.
As suggested by Lemma~\ref{lem:positional_unambiguity}, a view-dependent combined TSDF can be pre-computed by calculating the combined SDF for every voxel in the view frustum. This combined TSDF can then be used like a regular TSDF, but only for ray-casting from the pose used during combination. As a bonus, this opens up yet another class of tracking algorithms: the direct volume matching type, that perform registration directly within the SDF~\cite{Bylow2016, Canelhas2013, Slavcheva2018}.

To always use the most recent observations, all voxels that received new information during fusion also need to be updated in the combined TSDF. But for static scenes this is not always necessary.
Instead, we use conditional combination. Only meeting one of the following criteria triggers an update of the combined TSDF:
\begin{align}
    \mathrm{framesSinceStart} &< 5,&&\textit{boot up}\label{eq:condition_1} \\ 
    \mathrm{framesSinceLastUpdate} &> 50,&&\textit{stale state}\label{eq:condition_2}\\
    \lVert \mathrm{pose} - \mathrm{lastPose}\rVert_{\textrm{translation}} &> \SI{0.05}{m},&&\textit{translation}\label{eq:condition_3} \\
    \lVert \mathrm{pose} - \mathrm{lastPose}\rVert_{\textrm{angle}} &> 0.05 \frac{\pi}{2}.&&\textit{rotation}\label{eq:condition_4}
\end{align}
The boot up condition~\eqref{eq:condition_1} ensures, that during the first frames where the map is still uncertain, always the most recent data is used for tracking. \eqref{eq:condition_2} enforces regular updates in case the camera does not move. Eq.~(\ref{eq:condition_3},~\ref{eq:condition_4}) are a relaxation of Lemma~\ref{lem:positional_unambiguity}, that states minor changes in the camera pose don't change the combined TSDF, similar to small-motion assumption on which the data association for ICP is based~\cite{Newcombe2011}.
We experimentally chose the thresholds relatively small, so as not to violate the underlying assumption. A more thorough investigation on the impact of these limits would be interesting. On the tested sequences, the update is triggered on average around every third frame.
By also selecting voxels slightly beyond the camera frustum ($\pm 1/8$ image size), motions of the camera will not leave the scope of the combined TSDF before triggering a recalculation. Voxels that receive data for the first time are always directly added to the combined TSDF.

To prevent empty voxels in the absence of gradients in all directions (e.g., at edges), the weights
\begin{align}
  w^D_{\mathrm{noGrad}} = w^D \cdot \langle \mathbf{v}^D, -\mathbf{r} \rangle
  \label{eq:combine_tsdf_weight}
\end{align}
are used instead.

\section{ICP Tracking}
\label{sec:icp_tracking}
The geometric Iterative Closest Point algorithm optimizes the pose difference $\Delta T \in \mathbb{R}^{4\times 4}$ between two point clouds, or in our case a point cloud and a depth map. As introduced by Newcombe \etal~\cite{Newcombe2011}, the new depth frame is registered against a point cloud rendered from the previous pose estimate, starting with an estimate of $\Delta T = I_4$. This frame-to-render as compared to direct frame-to-model registration might seem to be an unnecessary intermediate step, but saves time on data lookups and the data is often already available, if the algorithm is running live with a visualization.
The algorithm uses projective data association to project each point from the depth frame at time $t+1$ into the frame rendered from the estimated pose at time $t$ according to the current estimate of $\Delta T$.

The rigid body transform $\Delta T$ is element of the Lie-group $\SE3$. Optimization takes place in the accompanying Lie-algebra $\se3$.
The 6-vector $\xi = (\nu, \omega) \in \se3$ with translation and rotation components $\nu, \omega \in \mathbb{R}^3$ is a minimal representation for the rigid body transform and the exponential map $\exp:\se3 \rightarrow \SE3$ converts from algebra to group elements. We adopt the notation from Blanco~\cite{Blanco2010}, who published a good review on the $\SE3$, including the most important derivatives. For improved readability, we omit conversions from and to homogeneous coordinates and use group elements of $\SE3$ synonymous with the respective transformation matrices $\in \mathbb{R}^{4\times 4}$.
Let $p_i \in \mathbb{R}^3$ be the i-th point of the depth frame and $q_i, n_i \in \mathbb{R}^3$ the associated point and normal in the rendered scene. Then the weighted geometric ICP formulation is
\begin{equation}
  E_\mathrm{geom} = \sum\limits_i w_i \underset{=:r_i}{\underbrace{\left\langle q_i - \exp(\xi)\Delta T p_i \mid n_i \right\rangle}}^2.
  \label{eq:geometric_ICP}
\end{equation}
The error is minimized using the Gauss-Newton method, where linearization around $\xi = 0$ allows to derive the Jacobian
\begin{equation}
  J_i = \left.\frac{\partial r_i}{\partial \xi}\right|_{\xi = 0}
  = \left( 
\begin{matrix}
  -n_i & -\Delta T p_i \times n_i
\end{matrix}
  \right)^\intercal \in \mathbb{R}^6
  \label{eq:ICP_geometric_dx}
\end{equation}

We had some serious issues with the repeatability of our experiments due atomic operations in the CUDA kernels: while computing the Hessian matrices during ICP with a three-level reduction pyramid, the resulting summands of the individual blocks get atomically added together. Floating point addition is, however, not commutative on computation hardware and we've seen deviation of our tracking results in the order of 1\%.
Therefore we changed summation to a fully-deterministic hierarchical scheme, with no noticeable increase in computation time.

\begin{align}
  H &= 2 \sum\limits_i w_i \sum\limits_{j, k} J_{i, j} J_{i, k} \in \mathbb{R}^{6\times 6}\\
  g &= 2 \sum\limits_i w_i J^\intercal_i r_i \in \mathbb{R}^6
  \label{eq:ICP_Hessian}
\end{align}
The equation system
\begin{equation}
  H \hat{\xi} = -g
  \label{eq:ICP_solution}
\end{equation}
can be solved efficiently using Cholesky decomposition, yielding solution $\hat{\xi}$ that is used to update the current pose estimate for the next iteration:
\begin{equation}
  \Delta T' = \exp(\hat{\xi}) \Delta T.
  \label{eq:deltaT_update}
\end{equation}
The algorithm iterates until the step size $\Vert\hat{\xi}\Vert_2$ falls below a certain threshold or a maximum number of iterations is reached.

\subsection{Weighted ICP}
\label{ssec:weighted_icp}
\begin{figure}[t]
  \centering
  \includegraphics[width=\linewidth]{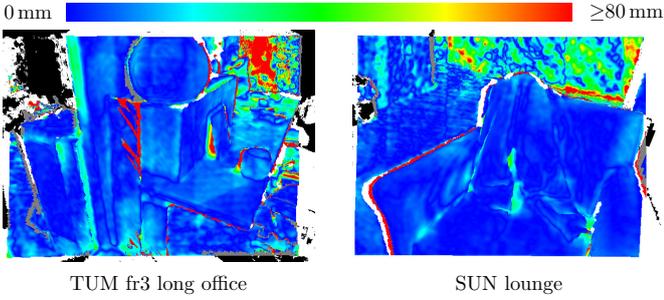}
  \caption{Depth noise examples, error increases from blue to red.}
  \label{fig:depth_error_examples}
\end{figure}
\begin{figure}[t]
  \centering
  \includegraphics[width=\linewidth]{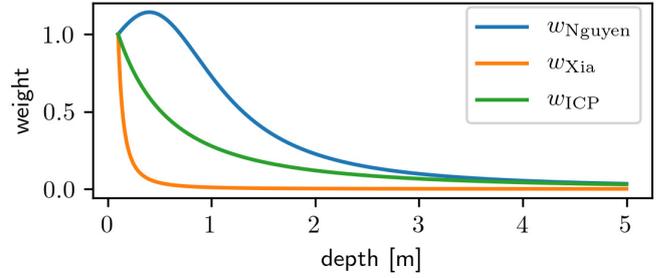}
  \setlength{\abovecaptionskip}{-10pt}
  \setlength{\belowcaptionskip}{-10pt}
  \caption{Visualization of depth-dependent ICP weights used in~\cite{Nguyen2012} ($w_\mathrm{Nguyen}$), \cite{Xia2018}~($w_\mathrm{Xia}$) and proposed ($w_\mathrm{ICP}$) for depth range $[0.1, 6]$ \SI{}{m}.}
  \label{fig:icp_weight_comparison}
\end{figure}

The error formulation in \eqref{eq:geometric_ICP} contains an optional per-pixel weight factor $w_i$. Surprisingly, while most TSDF implementations use some form of weighting during fusion, especially regarding the increasing unreliability with larger depth (c.f. \figref{fig:depth_error_examples}), only few implementations apply weights during tracking, when it is most crucial. Xia \etal~\cite{Xia2018} apply the following weight function for each depth value $z$ and valid depth range $z \in [z_\mathrm{min}, z_\mathrm{max}]$:
\begin{align}
  w_\mathrm{Xia} &= \frac{\frac{1}{z^2} - \frac{1}{z_\mathrm{max}^2}}{\frac{1}{z_\mathrm{min}^2} - \frac{1}{z_\mathrm{max}^2}}.
  \label{eq:ICP_weight_Xia}
\end{align}
Nguyen \etal~\cite{Nguyen2012} developed a more accurate, specific noise model for the Kinect v1 camera by measuring a flat panel at set distances and angles and fitting a function. They include an additional factor if the angle $\theta$ between surface normal to camera view ray exceeds 60\textdegree:
\begin{align}
  w_\mathrm{Nguyen} &= \frac{\sigma(z_\mathrm{min}, 0)}{\sigma(z, \theta)}, \\ 
  \sigma(z, \theta) &= 0.0012 + 0.0019 * (z - 0.4)^2 \left[ + \frac{0.0001}{\sqrt{z}}\frac{\theta^2}{\left( \frac{\pi}{2} - \theta \right)^2} \right].
  \label{eq:ICP_weight_Nguyen}
\end{align}
The summand in squared brackets is only applied for $\theta > 60^{\circ}$.
It is easy to spot that, apart from the angular factors, both weight functions use the inverse quadratic depth, but $w_\mathrm{Nguyen}$ has additional shaping parameters. As visualized in~\figref{fig:icp_weight_comparison} for a depth range of $[0.1, 6]$, the weight function $w_\mathrm{Xia}$ puts very high emphasis on close observations, so everything closer than \SI{0.5}{m} far outweighs other measurements. This creates problems in certain scenes, where objects sweep through the frame at a very close distance, e.g., it leads to tracking failure in the \emph{ICL deer walk} sequence.
With constant ICP weights, noisy sequences often fail completely (for instance, the noise-augmented sequences of the Zhou dataset, which have somewhat overexaggerated noise levels compared to real-world sensors).
While of course desirable, it is impractical to calibrate a specific noise model for every sensor. Therefore, we propose another depth weight, which has similar characteristics to the model by Nguyen \etal, but does not rely on any magic numbers:
\begin{align}
  w_\mathrm{ICP} = \frac{1}{\left(z + (1 - z_\mathrm{min})\right)^2}.
  \label{eq:ICP_weight}
\end{align}
The function stays within $[0, 1]$ and its asymptotic behavior still considers distant points to a reasonable degree.

As proposed before by Prisacariu \etal~\cite{Prisacariu2017}, we also performed experiments with the Huber loss as cost function, that theoretically should make tracking more robust against outliers, but found it to generally reduce tracking performance. Similar findings were reported by Bellekens \etal~\cite{Bellekens2015}. Therefore, we kept the standard quadratic error function.

\subsection{Photometric ICP}
\label{sec:photometric_icp}
Photometric ICP, that minimizes the photometric error between pixels of two RGB frames to determine the relative pose difference, has been used in RGB-D tracking for some time~\cite{Steinbruecker2011} and was successfully adapted to the TSDF as well~\cite{Whelan2015, Henry2013, Niesner2013}.

Setting aside the discussion of how to incorporate RGB data into tracking, a more fundamental question is which data to actually compare.
While geometric ICP relies on rendering a point cloud from the TSDF, for photometric ICP there are two directions throughout literature: frame-to-frame (\ftof) tracking~\cite{Whelan2015, Prisacariu2017}, where the frame at time $t$ is registered against the frame at time $t-1$, and frame-to-render (\ftor)~\cite{Niesner2013, Henry2013, Xia2018}, where the current frame is tracked against a point cloud rendered from the TSDF. The authors of~\cite{Bylow2016, Palazzolo2019} perform frame-to-model tracking, but the same discussion as above holds.
Frame-to-render has the advantage of tracking against a consistent model, thus it should be more robust against drift. On the other hand, because of the way colors are fused into the TSDF (without other augmentations), the contained photometric information is subject to blur, lighting and exposure changes, error accumulation etc. Moreover, the detail and sharpness directly correlate to the voxel size. Therefore, it would be plausible, that with increasing voxel sizes, the frame-to-render tracking deteriorates.
Frame-to-frame tracking, on the other hand, has the advantage of retaining the resolution of the input images, which for modern sensors (e.g., Intel RealSense) usually exceeds the depth image resolution. Whelan \etal~\cite{Whelan2015} argue, that very distant points that are useful for rotational constraining are usually not represented in the TSDF and also the TSDF resolution limits the effectiveness, though this argument only holds to a certain extent, as data association is performed by projecting depth points from the current image pair to the previous color image, so it is still bounded by the maximum perceivable depth of the sensor.
A disadvantage, besides the obvious drift, is the individual image quality which is subject to blur, rolling shutter, and depth-color synchronization issues.

Interestingly, to our knowledge, frame-to-keyframe (\ftokf) tracking, where the current frame is registered against a selected keyframe from the past, has not been applied in photometric tracking with the TSDF yet. It has been successfully implemented in model-less graph-based algorithms~\cite{Kerl2013,Mur-Artal2015,Dai2017}, effectively reducing drift, so we want to investigate its use in combination with the TSDF. We discuss this approach more in Section~\ref{ssec:keyframe_ICP}.

\begin{figure*}[h]
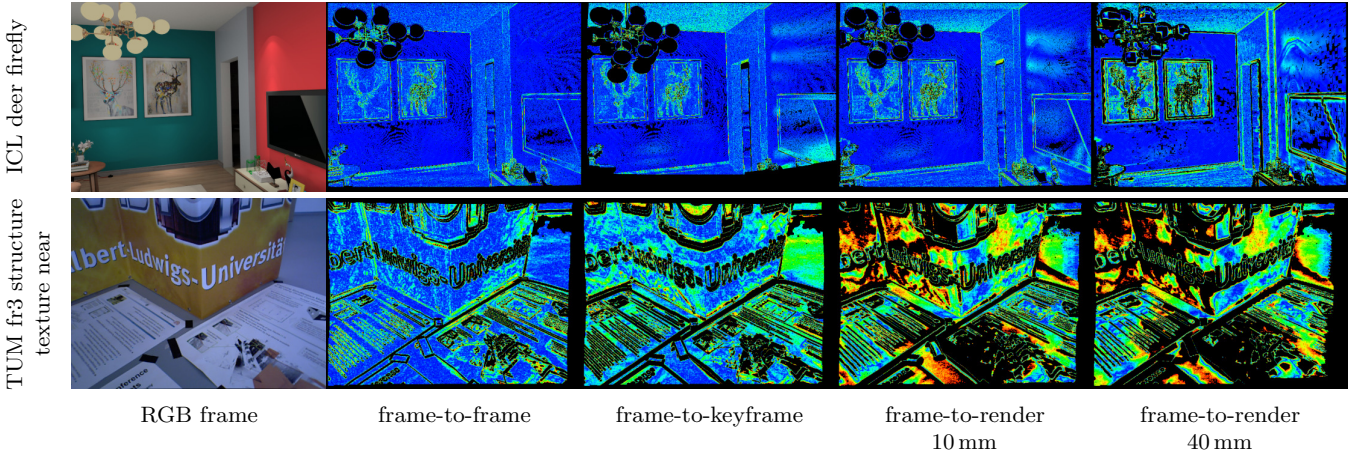

  \centering
  \resizebox{\linewidth}{!}{%
  \begin{tikzpicture}
    \node[node distance=3.6cm] (1) {\includegraphics[width=3.6cm]{images/photometric_error/ICL_rgb_0338.png}};
    \node[node distance=3.6cm, right of=1] (2) {\includegraphics[width=3.6cm]{images/photometric_error/ICL_f2f_0338.png}};
    \node[node distance=3.6cm, right of=2] (3) {\includegraphics[width=3.6cm]{images/photometric_error/ICL_f2kf_0338.png}};
    \node[node distance=3.6cm, right of=3] (4) {\includegraphics[width=3.6cm]{images/photometric_error/ICL_f2r_10_0338.png}};
    \node[node distance=3.6cm, right of=4] (5) {\includegraphics[width=3.6cm]{images/photometric_error/ICL_f2r_40_0338.png}};
    \node[node distance=2.8cm, below of=1] (21) {\includegraphics[width=3.6cm]{images/photometric_error/TUM_rgb_0290.png}};
    \node[node distance=3.6cm, right of=21] (22) {\includegraphics[width=3.6cm]{images/photometric_error/TUM_f2f_0290.png}};
    \node[node distance=3.6cm, right of=22] (23) {\includegraphics[width=3.6cm]{images/photometric_error/TUM_f2kf_0290.png}};
    \node[node distance=3.6cm, right of=23] (24) {\includegraphics[width=3.6cm]{images/photometric_error/TUM_f2r_10_0290.png}};
    \node[node distance=3.6cm, right of=24] (25) {\includegraphics[width=3.6cm]{images/photometric_error/TUM_f2r_40_0290.png}};
    \node[node distance=1.9cm, anchor=north, text width=4cm, below of=21, align=center] {{RGB frame\\~}};
    \node[node distance=1.9cm, anchor=north, text width=4cm, below of=22, align=center] {{frame-to-frame\\~}};
    \node[node distance=1.9cm, anchor=north, text width=4cm, below of=23, align=center] {{frame-to-keyframe\\~}};
    \node[node distance=1.9cm, anchor=north, text width=4cm, below of=24, align=center] {{frame-to-render\\\SI{10}{mm}}};
    \node[node distance=1.9cm, anchor=north, text width=4cm, below of=25, align=center] {{frame-to-render\\\SI{40}{mm}}};
    \node[node distance=2.0cm, text width=3cm, left of=1,rotate=90,align=center,anchor=south] {{ICL deer firefly\\~}};
    \node[node distance=2.0cm, text width=4cm, left of=21,rotate=90,align=center,anchor=south] {{TUM fr3 structure\\texture near}};
  \end{tikzpicture}
}
  \caption{Photometric error comparison for different reference image types for photometric ICP. Photometric error from blue (no error) to red (intensity threshold).
  }
  \label{fig:color_mode_error_comparison}
\end{figure*}
\figref{fig:color_mode_error_comparison} displays a comparison of photometric error for the different reference methods. All error images are rendered from the same ground truth poses.
Even given perfect poses from a noise-less artificial dataset (ICL deer firefly), the \ftof~and \ftokf~still show some error: a point from the current depth frame gets back projected into the reference color frame for data association and the photometric value is bilinearly interpolated, as the coordinates are not integer. The \ftokf~images (third column) have some missing pixels, as the warped keyframe is partially outside the current view.
The second row contains footage from the real-world dataset TUM fr3. Especially for larger voxel sizes where many colors, lighting conditions and image exposures blend together, the rendered color image are very blurred. Therefore, many pixels fall below the minimum gradient threshold or above the maximum intensity threshold that prevent using pixels with unreliable information during optimization.

Only using photometric ICP does not provide reliable tracking, so we use combined ICP as described in~\cite{Whelan2015, Xia2018}, where the error functions are combined as 
\begin{equation}
  E = E_\mathrm{geom} + \lambda E_\mathrm{photo}.
  \label{eq:combined_icp}
\end{equation}
Most combined ICP approaches use a fixed weight between photometric and geometric terms for the entire image pair, though its role is not clearly stated in all works: it is required to compensate the difference in metric between depth and intensity errors. A reoccurring constant weight throughout literature is 0.1, though there are some discrepancies whether the weight is squared and whether the kernel for gradient computation is normalized~\cite{Henry2013, Whelan2015, Xia2018}.
We used 0.1 without squaring and, upon inspection, the magnitudes of the geometric and photometric Hessian matrices were in the same order.

In their implementation Whelan \etal~\cite{Whelan2015} additionally use the root mean square of the intensity difference to down-weight pixels with larger errors, but didn't find any good explanation. We apply the same per-pixel depth weight as before, as the data association of color pixels also depends on the quality of the depth.

\subsection{Keyframe Photometric ICP}
\label{ssec:keyframe_ICP}
For combined ICP, The frame-to-frame and frame-to-render formulations use the fact that the model is rendered at the same location as the RGB reference image frame. With frame-to-keyframe ICP, the pose of the reference (key)frame at time $t^\mathrm{kf}$ differs from the pose of the rendered scene at time $t$. Let $T^\mathrm{RS}$ be the transform from rendered scene to reference frame and $\Delta T$, just like above, the pose estimate from rendered scene to frame $t+1$ of the current ICP iteration (initialized with $I_4$). Then the error formulation becomes
\begin{equation}
\resizebox{.89\hsize}{!}{$
  E_\mathrm{RGB}^{\mathrm{kf}}
  = \sum\limits_i \left\lVert
 \underset{=:r(\xi, p_i)}{\underbrace{
   I^{t+1}\left(\Pi(p_i)\right) - I^{t^\mathrm{kf}}\left( \Pi \left( K \cdot T^{\mathrm{RS}}\exp(\xi) \Delta T \cdot p_i \right) \right)}}
  \right\rVert^2.
$}
  \label{eq:keyframe_icp}
\end{equation}
with camera matrix $K$, projective function $\Pi$ (\eqref{eq:projection_dx}). Linearization around $\xi = 0$ gives
\begin{equation}
  \frac{\partial r}{\partial \xi} (\xi, p_i)
  \approx
  r(0, p_i) + \frac{\partial r}{\partial \xi} (0, p_i) \cdot \xi
  \label{eq:keyframe_icp_derivative}
\end{equation}
with the derivative being
\begin{align}
  &\frac{\partial r}{\partial \xi} (0, p_i) \nonumber \\
  =&\mbox{$\frac{\partial I^\mathrm{kf}}{\partial \Pi} \cdot \frac{\partial \Pi}{\partial K} \cdot \frac{\partial K}{\partial p_i} \cdot \frac{\partial T^\mathrm{RS} \exp(\xi) \Delta T p_i}{\partial \xi}$} \\
  =&\mbox{$\nabla I^\mathrm{kf} \cdot \left. \frac{\partial \Pi}{\partial a} \right|_{a = K T^\mathrm{RS} \Delta T p_i} \cdot K \cdot R(T^\mathrm{RS})
  \left[ I_3 \mid -(\Delta T p_i)^{\wedge} \right]
.$}
  \label{eq:keyframe_icp_derivativea_2}
\end{align}
$R(\cdot)$ extracts the rotation matrix from the transform. The wedge operator for a vector $\nu = (x, y, z)^\intercal \in \mathbb{R}^3$ is defined as
\begin{equation}
  \nu^\wedge = 
  \left( 
  \begin{matrix}
    0 & -z & y \\
    z & 0 & -x \\
    -y & x & 0
  \end{matrix}
  \right).
  \label{eq:wedge_operator}
\end{equation}
The derivative of the projection function is
\begin{equation}
  \frac{\partial \Pi}{\partial (x, y, z)^\intercal} =
  \frac{\partial}{\partial (x, y, z)^\intercal} 
  \left[
    \begin{matrix}
      \frac{x}{z} \\
      \frac{y}{z}
    \end{matrix}
  \right]
  = \left[ 
    \begin{matrix}
      \frac{1}{z} & 0 & \frac{x}{z^2} \\
      0 & \frac{1}{z} & - \frac{y}{z^2}
    \end{matrix}
  \right].
  \label{eq:projection_dx}
\end{equation}
For the \ftof~scenario, the same formulation can be used with $T^\mathrm{RS} = I_4$.

\paragraph{Keyframe Selection}
The choice of keyframes is the problem of finding a compromise between good coverage of the scene and using only as few views as necessary. Most approaches in the past used heuristics of different complexity to find suitable keyframes.
The easiest way is to select a new keyframe at fixed intervals, as it is done in ORB-SLAM~\cite{Mur-Artal2015}. Other common heuristics are thresholding the change of angular and translational change~\cite{Engel2014, Forster2017, Engel2018}. Sucar \etal~\cite{Sucar2021} compute a measure on how much of the current depth image is not yet represented by the map. Some methods also decide based on image quality, by selecting frames with low jitter, camera velocity~\cite{Yang2017} or when the exposure time changed too much~\cite{Engel2018}. The downside to these hand-crafted metrics is, that they require tuning of multiple parameters, possibly even depending on the scene.
Entropy based selection schemes use the covariance of the Hessian matrix of the Gauss-Newton algorithm to determine when a new keyframe is due. Kerl \etal~\cite{Kerl2013} compare the negative entropy of the keyframe and the current frame and if the ratio falls below a threshold, select a new keyframe. Because the average negative entropy levels can drift with the camera moving (especially they can also increase), Kuo \etal~\cite{Kuo2020} apply a running average filter to the entropies for comparing.
The main advantage of this method is that this only leaves one threshold as tunable parameter. In practice we found, though, that tuning these methods is not that simple. High frequency fluctuations can cause fast change of keyframes, but using a more conservative threshold often results in keeping the keyframe for too long. Because of the running average adapting to the overall trend of negative entropies, the method proposed by Kuo~\etal~\cite{Kuo2020} would sometimes not select any new keyframes at all.

In our experiments we want to investigate, whether using keyframes can help mitigate drift and jitter. For comparability, it is important that all runs use the same keyframes, as the heuristics described above might select different keyframes based on the differences in tracking. Therefore, we chose a very simple scheme: every 10 frames a new keyframe is selected. All datasets used in our experiments have an image rate of 20-\SI{30}{Hz}, so this allows for large enough overlap, but is far enough apart to tell if there is any advantage or disadvantage compared to frame-to-frame tracking.

\section{Multi-Sensor Sim3 Scale ICP}
\label{sec:sim3_icp}
\begin{figure*}[t]
  \hspace*{\fill}%
  \begin{subfigure}[b]{.45\linewidth}
    \includegraphics[width=\linewidth]{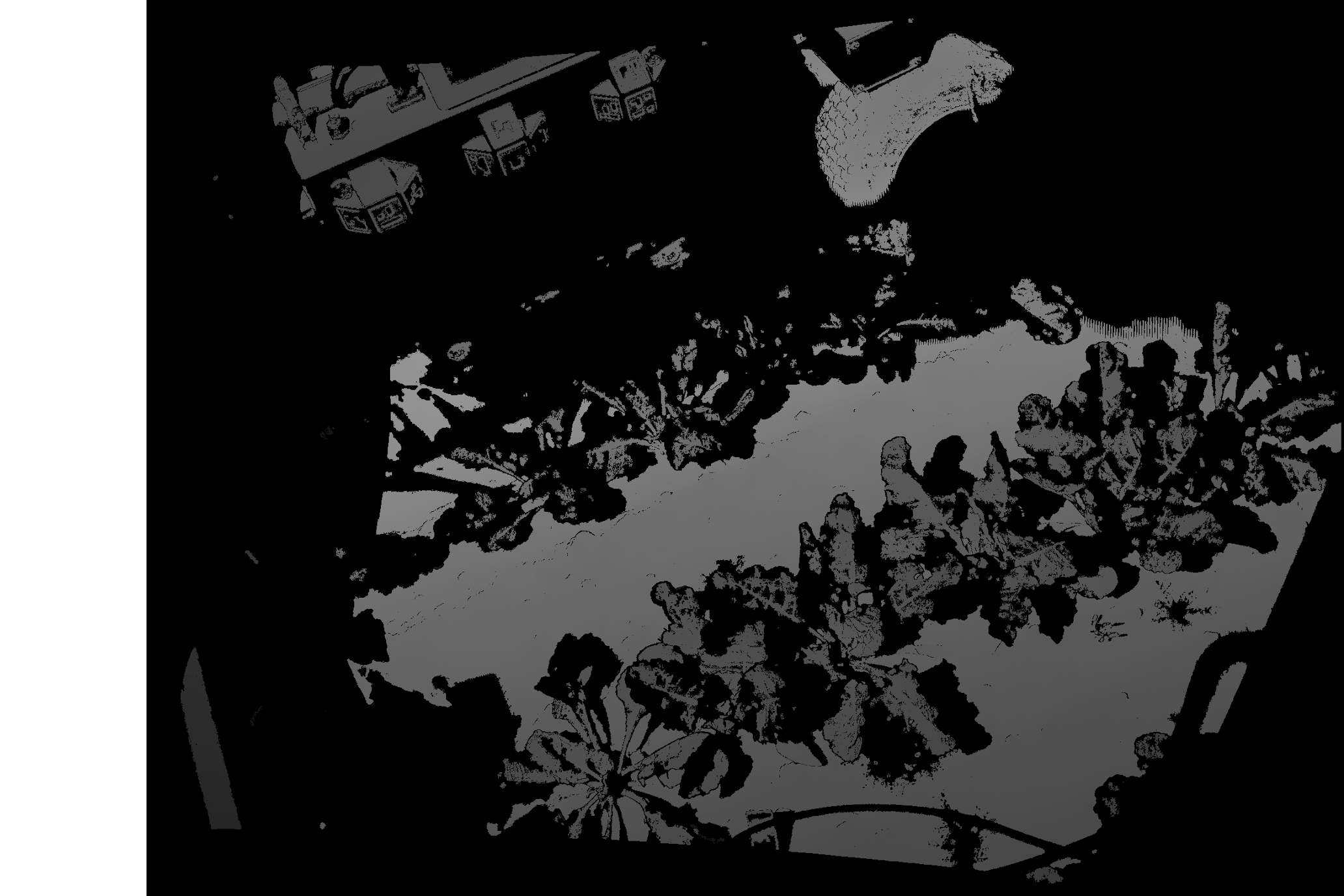}
    \caption{Photoneo depth}
  \end{subfigure}
  \begin{subfigure}[b]{.45\linewidth}
    \includegraphics[width=\linewidth]{images/stereo_rgb_example.png}
    \caption{DSLR RGB image}
  \end{subfigure}
  \hspace*{\fill}%
  \caption{Example data from PhenoRob UGV dataset.}
  \label{fig:phenorob_data_example}
\end{figure*}

Part of our work in the PhenoRob Cluster of Excellence is to reconstruct crops from data collected on the field by a UGV. The UGV is an inverse U-shape constructed from aluminum struts, driving over the crops and taking a snapshot with 14 Nikon DSLR cameras (five stereo pairs, four monocular cameras) and five Photoneo PhoXi 3D scanners, producing high-accuracy depth maps. The robot stops over a crop patch, takes a snapshot with all sensors and then proceeds to the next patch. \figref{fig:phenorob_data_example} shows sample plant measurements.

One challenge is to fuse the data from different sensors, as there are slight depth discrepancies. Moreover, in this scenario the extrinsic parameters of the cameras change when the robot moves, because of its size and limited rigidity. Also, the stereo-pair baseline is used to convert disparity images to depth maps, so the depth might vary, based on small changes in the camera setup. To compensate for this while fusing the data, we extend the standard $\SE3$ ICP tracking, which estimates translation and rotation, to the similarity transform $\Sim3$, which jointly optimizes pose and scale. We apply this method to refine the depth scale and pose priors provided by the extrinsic calibration of the multiple sensors.

For the point-to-point scale ICP least-squares optimization, there exists a closed form solution of rotation, translation and scale, first described by Horn~\cite{Horn1987}. This method is, however, not applicable for partially similar point clouds. This problem was addressed by Sahillioglu \etal~\cite{Sahillioglu2021}. TEASER~\cite{Yang2020} is an algorithm that is highly robust to large amounts of outliers and non-zero-mean Gaussian noise.
Du \etal~\cite{Du2010} proposed an algorithm that optimizes the scales for each coordinate axis. LSD-SLAM~\cite{Engel2014} performs $\Sim3$ photometric ICP to unify the scale-uncertainty of monocular SLAM.

Chen and Medioni~\cite{Chen1992} state, that with unknown correspondences ICP with the point-to-plane metric converges quicker than point-to-point, so it is the more sensible choice in our case. Note, that without one-to-one point correspondences this is an ill-posed problem, as translation and scale optimization to a degree modify the same error gradient (in z-direction) and because of the projective data association, especially for environments with little geometric diversity like corridors etc., for a majority of points both parameters get optimized. In other words there are multiple solutions with similar low residuals, but different bias towards scale and translation optimization. In the extreme case of, for instance, perceiving only a plain a wall, optimizing either pose or scale will have identical results. 

We use the same error formulations \eqref{eq:geometric_ICP} and \eqref{eq:keyframe_icp} as before, only with $\xi = (\nu, \omega, \sigma)^\intercal \in \sim3$ instead. The additional 7th parameter represents the scale and the corresponding exponential map is
\begin{align}
  \exp:\qquad\sim3 &\rightarrow \Sim3,\\
  (\nu, \omega, \sigma)^\intercal &\mapsto
  \left[
  \begin{matrix}
    \exp(\sigma) \exp(\omega) & V \nu \\
    0 & 1
  \end{matrix}
  \right].
  \label{}
\end{align}
For $A, D \in \Sim3, p \in \mathbb{R}^3$ we require the derivative of the transposed point at $\xi = 0$:
\begin{equation}
  \left. \frac{\partial}{\partial \xi} A \oplus \exp(\xi) \oplus D \oplus p \right|_{\xi = 0} =
  R(A)
  \left[ 
    \begin{matrix}
      I_3 & -(D \oplus p)^{\wedge} & D \oplus p
    \end{matrix}
  \right]
  \label{eq:sim3_derivative}
\end{equation}
\eqref{eq:sim3_derivative} can be directly inserted into \eqref{eq:ICP_geometric_dx} and \eqref{eq:keyframe_icp_derivativea_2} to get the respective Jacobians. The rest of the algorithm works analogously, with the minor difference, that the resulting Hessian is $7 \times 7$ instead of $6 \times 6$.

For camera intrinsics $(f_x, f_y, c_x, c_y)$, image coordinates $x, y$, and depth value $d$, the depth reprojection function~\eqref{eq:reprojection} is linear in $d$, so scaling the depth is equivalent to scaling the point in the camera frame:
\begin{equation}
  \Pi^{-1}(x, y, d) = \left( d\frac{x - c_x}{f_x}, d \frac{y - c_y}{f_y}, d \right)^T.
  \label{eq:reprojection}
\end{equation}
This implies that instead of changing the entire TSDF pipeline to use $\Sim3$ poses, it is possible to just scale the input images with the given factor and use the corresponding $\SE3$ poses for everything else.

\begin{figure*}[t]
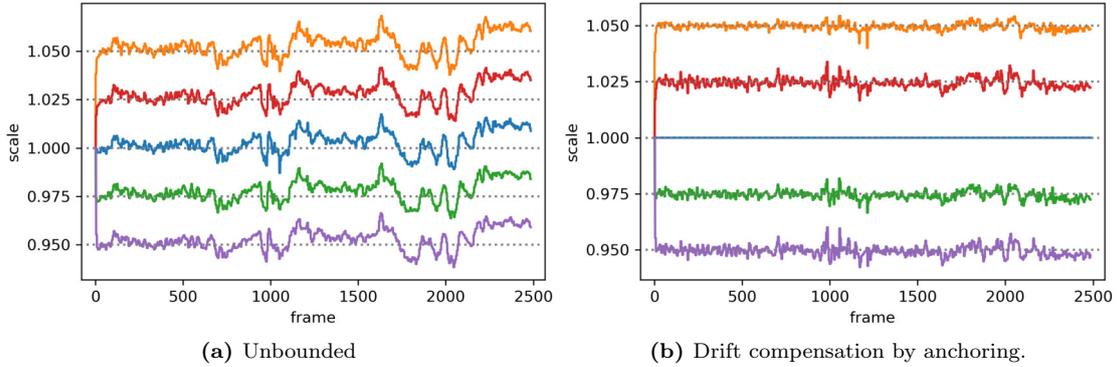

  \hspace*{\fill}%
  \begin{subfigure}[b]{0.40\linewidth}
    \centering
    \includegraphics[width=\linewidth]{images/scale_multi_sensor_no_anchor.png}
    \caption{Unbounded}
    \label{fig:scale_multi_sensor_no_anchor}
  \end{subfigure}
  \begin{subfigure}[b]{0.40\linewidth}
    \centering
    \includegraphics[width=\linewidth]{images/scale_multi_sensor_compensate.png}
    \caption{Drift compensation by anchoring.}
    \label{fig:scale_multi_sensor_compensate}
  \end{subfigure}
  \hspace*{\fill}%
  \caption{Comparison of unbounded- and anchored scale estimation for five sensors. Gray dotted lines are the true scale factors. The colored lines correspond to the scale factors estimated for the individual sensors; the anchored sensor in the (\subref{fig:scale_multi_sensor_compensate}) is represented by the blue line.}
  \label{fig:scale_multi_sensor}
\end{figure*}

To test the behavior in a controlled manner, we transform the \emph{fr3\_long\_office} sequence into an artificial five-sensor sequence, by scaling batches of five consecutive depth images with constant factors $(1.0, 1.05, 0.975, 1.025, 0.95)$, such that the sequence consists of a repeated cycle of five different sensors.
Just like with pose tracking, estimation of scale against the model accumulates drift over time, as \figref{fig:scale_multi_sensor_no_anchor} shows. Noticeably, the drift happens in unison across the sensors. By anchoring the scale of one of the sensors to a fixed value $\hat{s_0}$ and compensating the other estimates, the drift is effectively prevented and absolute error of scale estimates is significantly lower, as can be seen in~\figref{fig:scale_multi_sensor_compensate}. Let $\hat{s_i}$ be the estimated scale of sensor $i > 0$ at time $t$. Then the compensated scale is computed as
\begin{equation}
  s_i^t = \exp(\hat{s_i}^t) \exp(\hat{s_0})^{-1} = \hat{s_i}^t - \hat{s_0}.
  \label{eq:scale_anchor}
\end{equation}

\begin{figure*}[p]
  \includegraphics[width=\linewidth, compress=false]{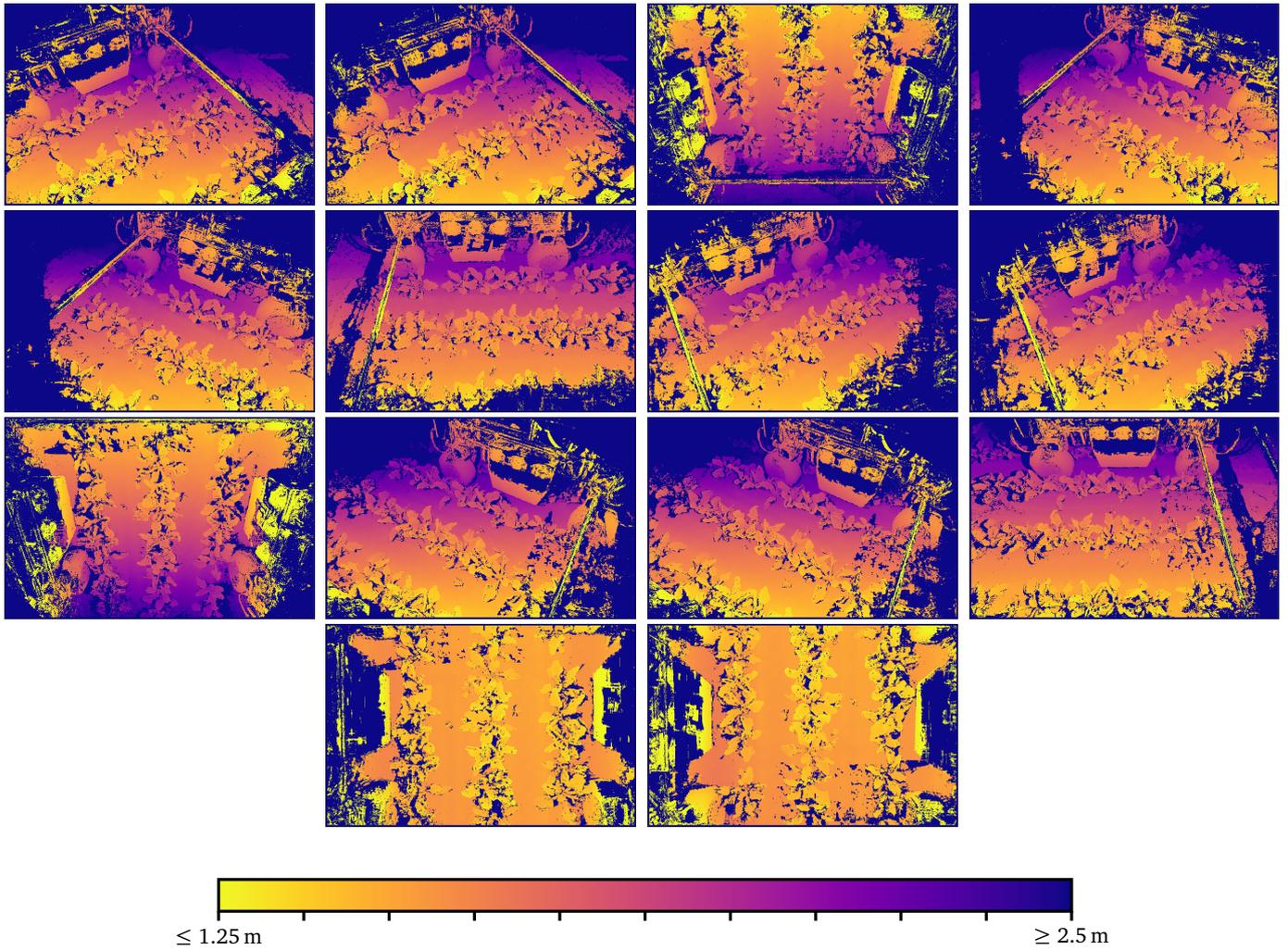} 
  \caption{Example depth images generated by COLMAP.}
  \label{fig:colmap_output}
\end{figure*}

\figref{fig:phenorob_data_example} depicts examples of typical data acquired by the robot on the field. Note, how the Photoneo sensor yields many missing measurements around the plants. These are caused by occlusions and wind moving the plants during scans, as the acquisition process takes 250 -- \SI{2500}{ms}.
We use Kalibr~\cite{Furgale2013} to determine intrinsic and extrinsic parameters of the RGB cameras and multiple ArUco markers on the side panels of the robot to align RGB and the Photoneo sensors. For generating depth from stereo, we have experimented with classical methods like Semi-Global Matching~\cite{Hirschmueller2005} and the hierarchical deep stereo matching method by Yang \etal~\cite{Yang2019}, but have not yet generated satisfying results, i.e., the depth error is too large to reliably overlay multiple views. %
Instead, we use the multi-view stereo (MVS) pipeline COLMAP~\cite{Schoenberger2016,Schoenberger2016a} to generate cross-matched depth maps for all RGB cameras. Owing to the large overlap of camera views, a depth map is generated for each of the 14 cameras. \figref{fig:colmap_output} shows a set of all 14 depth images that form a single scan.

\begin{figure*}[p]
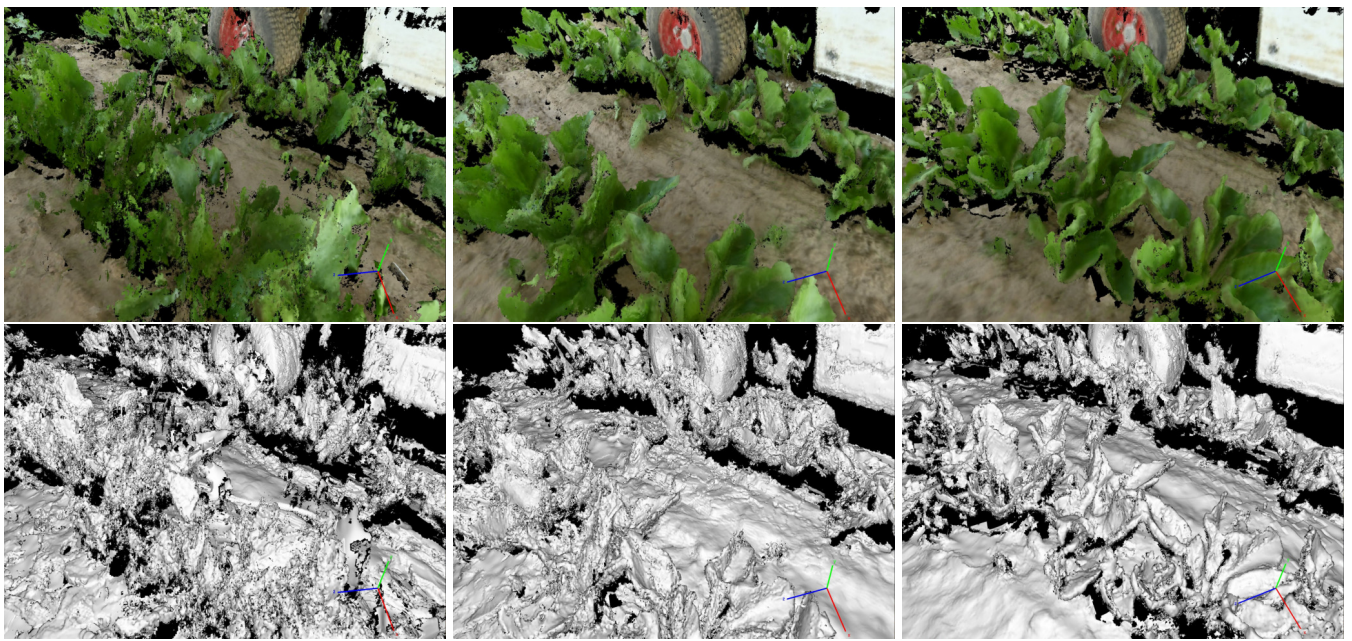

  \centering
  \begin{subfigure}[b]{.32\linewidth}
    \includegraphics[width=\linewidth]{images/phenorob/color_dtsdf_no_align.png} 
  \end{subfigure}
  \begin{subfigure}[b]{.32\linewidth}
    \includegraphics[width=\linewidth]{images/phenorob/color_dtsdf_se3.png} 
  \end{subfigure}
  \begin{subfigure}[b]{.32\linewidth}
    \includegraphics[width=\linewidth]{images/phenorob/color_dtsdf_sim3.png} 
  \end{subfigure}
  \begin{subfigure}[b]{.32\linewidth}
    \includegraphics[width=\linewidth]{images/phenorob/geometry_dtsdf_no_align.png} 
    \caption{No pose refinement.}
  \end{subfigure}
  \begin{subfigure}[b]{.32\linewidth}
    \includegraphics[width=\linewidth]{images/phenorob/geometry_dtsdf_se3.png} 
    \caption{$\SE3$ pose refinement.}
  \end{subfigure}
  \begin{subfigure}[b]{.32\linewidth}
    \includegraphics[width=\linewidth]{images/phenorob/geometry_dtsdf_sim3.png} 
    \caption{$\Sim3$ pose refinement.}
  \end{subfigure}
  \caption{PhenoRob UGV sugar beet reconstructions with and without pose refinement. Voxel size \SI{2.5}{mm}.}
  \label{fig:phenorob_reconstruction_color}
\end{figure*}
While COLMAP already computes good poses, which don't require further refinement, MVS does not produce an absolute metric scale, so we use the depth of the high-accuracy Photoneo PhoXi sensors and register the MVS depth maps against them.
The initial guess of the optimization procedure takes as input the scale estimate which is visually close to the Photoneo depth, and use the extrinsics from our offline calibration process, including the RGB sensor poses, so all sensors get registered.
\figref{fig:phenorob_reconstruction_color} shows example reconstructions without alignment, with $\SE3$ and with $\Sim3$ ICP pose refinement in comparison. $\Sim3$ optimization produces the smallest error.

\section{Implementation Details}
\label{sec:implementation_details}
Our implementation is based on InfiniTAM~\cite{Kaehler2015}, with significant modifications. For optimized memory usage, the voxel hashing scheme introduced by Nießner \etal~\cite{Niesner2013} is used. Voxels are allocated in blocks of $8\times 8 \times 8$ only where required and a hash map is used for constant-time access. Among other changes, the implementation now uses the stdgpu library by Stotko~\cite{Stotko2019} to replace several components, especially the original hash map, which could not allocate blocks with colliding hash values within the same iteration.

Unlike the previous DTSDF implementation presented in~\cite{Splietker2019} where 6 separate TSDF volumes were used for the different directions, here only a single TSDF is utilized. The hash index is extended from $(x,y,z) \in \mathbb{Z}^3$ to $(x, y, z, D) \in \mathbb{Z}^3 \times \mathrm{Directions}$. %
This simplifies many functions and better utilizes the statically allocated memory on the GPU: For most scenes the DTSDF has an imbalance of direction-usage, which wastes a lot of memory in the old scheme. Resizing the volumes is an option, but requires additional overhead which can be simply avoided by the aforementioned modification.

As a proof of concept, we use the pipeline depicted in \figref{fig:iconic} with the geometric and combined ICP tracker described in \secref{sec:icp_tracking}. We solve the optimization with the Levenberg-Marquardt (LM) algorithm. The originally implemented LM damping scheme, where the damping factor is multiplied/divided by 10 whenever the error increases/decreases was replaced by the scheme proposed by Madsen \etal~\cite{Madsen2004}, which promises better convergence while avoiding premature convergence towards local minima.

The time for computing the rendering TSDF is crucial for real-time usage of our method. It can be significantly sped up by taking advantage of the lookup positions being only integer voxel positions. Hence, no trilinear interpolation is required and the gradient can be computed with just looking up the SDF values stored in the 6 neighboring voxels. We further managed to halve the time by pre-caching the TSDF lookups for all voxels of the same block in shared memory. At the very most, for every direction there are the current block and its six neighboring blocks, so a total of 48 are looked up at the beginning.

Image preprocessing includes a depth filter. Especially the artificial datasets with noise augmentation contain an abundance of noisy pixels around object boundaries. If there are not at least two depth pixels in the direct neighborhood, that support the depth value, the pixel is discarded. Depth normals are approximated as cross product of neighboring pixels in x- and y direction and afterwards processed by a bilateral filter.

\section{Evaluation}
\label{sec:evaluation}
The datasets used in our evaluation are the Stanford 3D Scene Data (totempole, etc.)~\cite{Zhou2013}, ICL NUIM~\cite{Handa2014} (\emph{lr} and \emph{office}), Zhou~\cite{Choi2015}, the TUM RGB-D benchmark~\cite{Sturm2012} (\emph{fr1} and \emph{fr3}) as well as (new) ICL~\cite{Saeedi2019} (\emph{deer} and \emph{diamond}). For improved readability, we omit the dataset name where possible.
Frame-to-frame, frame-to-keyframe, and frame-to-render are abbreviated as \ftof, \ftokf~and \ftor, respectively. SotA refers to the state-of-the-art regular TSDF.

Our evaluation compares and analyzes the impact of selected parameters on the results of the algorithm, which will be clearly stated for the experiments. All other parameters are consistent throughout all runs and were chosen based on experiments and values found in related work to give a good balance between convergence reliability and tracking quality:
\begin{itemize}
  \item depth outlier filter as described in \secref{sec:implementation_details},
  \item bilateral depth filter ($\sigma_d = 5.0, \sigma_r = 0.025, \mathrm{radius} = 5$), 
  \item bilateral normal filter ($\sigma_d = 2.5, \sigma_r = 5.0, \mathrm{radius} = 5$),
  \item ICP settings
    \begin{itemize}
      \item termination condition minimum step size $10^{-6}$\SI{}{m},
      \item iteration upper bound 20 coarse / 50 fine,
      \item depth outliers threshold \SI{0.05}{m} coarse / \SI{0.005}{m} fine and
      \item intensity outlier threshold 0.175 coarse / 0.05 fine.
    \end{itemize}
\end{itemize}

\subsection{ICP Tracking}
For comparing the tracking performance, we evaluate the regular TSDF (marked state-of-the-art, SotA) against the DTSDF by running scenes from the aforementioned datasets and comparing the tracking results against the provided ground-truth trajectory using the relative pose error (RPE) with a window size of 30 frames (\SI{1}{s}). Note that this study does not try to compare to complete SLAM algorithm with loop closure detection and correction, but showcases the performance of the DTSDF relative to the regular TSDF as an enhanced data structure. All settings are equal across both modes and the tracker uses the default geometric ICP algorithm.

As expected, tracking does benefit from the DTSDF in scenes, where the camera observes thin structure from different angles. Otherwise, there is no significant difference.

The first test on artificially generated sequences from the ICL NUIM and Zhou datasets is reported in \tabref{tab:tracking_ICL} for different voxel sizes. Note that the noise-augmented sequences are being used.
The tracking performance is similar for most sequences, which is likely due to the mapped environments, which are convex rooms where the regular TSDF does not display its issues.
The Zhou office sequences, a scene of cluttered office desks scanned from different directions, provides an environment where the DTSDF actually has an advantage, which is reflected in the RPE.
\tabref{tab:tracking_models} does the same comparison on the turntable-like dataset used in the original DTSDF paper~\cite{Splietker2019}. In those sequences the camera orbits around a center point and only the model is visible, which is challenging to track due to the details and high percentage of thin structure w.r.t. the whole scene. The RPE distinctly shows the strength of the DTSDF.
\tabref{tab:tracking_tum} shows the results with real-world scans from the TUM dataset, with very similar results. Large planar surfaces with sharp corners (structure notex sequences) seem to benefit from the DTSDF. Here, the measurements have to be considered with care, as the underlying ground-truth is not perfect. 

\newcolumntype{C}[1]{>{\centering\arraybackslash}p{#1}}
\newcolumntype{R}[1]{>{\raggedleft\arraybackslash}p{#1}}
\begin{table}[h]
  \caption{Tracking RPE in \SI{}{mm}, mean memory usage, and per-frame runtime of synthetic ICL NUIM~\cite{Handa2014} and Zhou~\cite{Choi2015} sequences for different voxel sizes.}
  \label{tab:tracking_ICL}
  \vspace{-5mm}
  \center
\resizebox{\linewidth}{!}{
  \begin{tabular}{p{18mm}@{\extracolsep{4pt}}R{8mm}R{10mm}R{8mm}R{10mm} R{8mm}R{10mm}}
    voxel size & \multicolumn{2}{c }{5}& \multicolumn{2}{c }{10}& \multicolumn{2}{c}{20} \\
    \cline{2-3} \cline{4-5} \cline{6-7}
    & SotA & DTSDF & SotA & DTSDF & SotA & DTSDF\\
    \hline
    lr kt0n & 10.5 & \textbf{10.3} & 9.4 & \textbf{9.3} & \textbf{9.5} & \textbf{9.5} \\
    lr kt1n & \textbf{9.5} & \textbf{9.5} & 9.9 & \textbf{9.6} & 9.7 & \textbf{9.4} \\
    lr kt2n & \textbf{15.3} & \textbf{15.3} & 15.6 & \textbf{15.5} & \textbf{15.5} & \textbf{15.5} \\
    lr kt3n & 15.6 & \textbf{11.7} & 29.1 & \textbf{14.9} & \textbf{16.8} & 87.4 \\
    \hline
    office kt0n & \textbf{9.7} & \textbf{9.7} & \textbf{9.7} & \textbf{9.7} & \textbf{9.7} & 9.8 \\
    office kt1n & \textbf{9.5} & \textbf{9.5} & \textbf{9.5} & \textbf{9.5} & \textbf{9.5} & \textbf{9.5} \\
    office kt2n & \textbf{15.4} & 15.5 & \textbf{15.6} & 16.0 & \textbf{15.4} & \textbf{15.4} \\
    office kt3n & 11.0 & \textbf{10.9} & 11.3 & \textbf{10.9} & 11.2 & \textbf{11.0} \\
    \hline
    Zhou lr1 & 1.5 & \textbf{0.8} & 1.3 & \textbf{1.1} & 2.4 & \textbf{2.2} \\
    Zhou lr2 & 0.8 & \textbf{0.7} & 1.1 & \textbf{1.0} & 2.3 & \textbf{2.0} \\
    Zhou office1 & 1.2 & \textbf{1.0} & 3.2 & \textbf{1.3} & 10.5 & \textbf{3.1} \\
    Zhou office2 & 1.0 & \textbf{0.8} & 3.6 & \textbf{1.4} & 14.8 & \textbf{2.4} \\
    \hline \hline
    $\varnothing$ time [\SI{}{ms}] 
    & \textbf{8.4} & 10.7 & \textbf{6.4} & 7.1 & \textbf{5.7} & 6.0 \\
    \hline
    $\varnothing$ mem [\SI{}{MB}] 
    & \textbf{1350} & 1855 & \textbf{326} & 502 & \textbf{75} & 137
  \end{tabular}
}
\end{table}
\begin{table}[h]
  \caption{Tracking RPE in \SI{}{mm}, mean memory usage, and per-frame runtime of synthetic sequences rendered from Stanford 3D models~\cite{Splietker2019} for different voxel sizes.}
  \label{tab:tracking_models}
  \vspace{-5mm}
  \center
\resizebox{\linewidth}{!}{
  \begin{tabular}{p{18mm}@{\extracolsep{4pt}}R{8mm}R{10mm}R{8mm}R{10mm} R{8mm}R{10mm}}
    voxel size & \multicolumn{2}{c }{5}& \multicolumn{2}{c }{10}& \multicolumn{2}{c}{20} \\
    \cline{2-3} \cline{4-5} \cline{6-7}
    & SotA & DTSDF & SotA & DTSDF & SotA & DTSDF\\
    \hline
    armadillo & 2.8 & \textbf{2.7} & 2.9 & \textbf{2.5} & 10.6 & \textbf{5.6} \\
    bunny & \textbf{2.0} & \textbf{2.0} & 2.3 & \textbf{1.9} & 9.4 & \textbf{4.8} \\
    dragon & 3.0 & \textbf{2.9} & 4.7 & \textbf{3.4} & 21.6 & \textbf{10.6} \\
    turbine blade & 10.4 & \textbf{6.7} & 16.9 & \textbf{10.1} & 84.8 & \textbf{13.5} \\
    \hline
    $\varnothing$ time [\SI{}{ms}] 
    & \textbf{4.9} & \textbf{4.9} & 4.8 & \textbf{4.7} & \textbf{4.7} & \textbf{4.7} \\
    \hline
    $\varnothing$ mem [\SI{}{MB}] 
    & \textbf{18} & 41 & \textbf{4} & 13 & \textbf{1} & 4
  \end{tabular}
}
\end{table}
\begin{table}[h!]
  \caption{Tracking RPE in \SI{}{mm}, mean memory usage, and per-frame runtime of TUM sequences~\cite{Sturm2012} for different voxel sizes.}
  \label{tab:tracking_tum}
  \vspace{-5mm}
  \center
\resizebox{\linewidth}{!}{
  \begin{tabular}{p{18mm}@{\extracolsep{4pt}}R{8mm}R{10mm}R{8mm}R{10mm} R{8mm}R{10mm}}
    voxel size & \multicolumn{2}{c }{5}& \multicolumn{2}{c }{10}& \multicolumn{2}{c}{20} \\
    \cline{2-3} \cline{4-5} \cline{6-7}
    & SotA & DTSDF & SotA & DTSDF & SotA & DTSDF\\
    \hline
    desk1 & 62.9 & \textbf{59.5} & 63.7 & \textbf{58.0} & 66.6 & \textbf{60.6} \\
    long office & 24.3 & \textbf{24.1} & 26.8 & \textbf{25.2} & \textbf{25.5} & 25.8 \\
    \multirow{2}{*}{\shortstack[l]{structure\\notex far}} & 12.1 & \textbf{12.0} & 12.2 & \textbf{12.1} & 12.1 & \textbf{12.0} \\
    & & & & \\
    \multirow{2}*{\shortstack[l]{structure\\notex near}} & 15.3 & \textbf{15.1} & 15.3 & \textbf{15.2} & \textbf{15.4} & \textbf{15.4} \\\\
    \hline
    $\varnothing$ time [\SI{}{ms}] 
    & \textbf{7.0} & 10.3 & \textbf{5.9} & 6.6 & \textbf{5.6} & 5.9 \\
    \hline
    $\varnothing$ mem [\SI{}{MB}] 
    & \textbf{670} & 1419 & \textbf{132} & 332 & \textbf{28} & 84
  \end{tabular}
}
\end{table}

\subsection{Map Reusability}
Overall, the tracking results show that the DTSDF generally only has an advantage in sequences, where the camera observes structure from different sides. In those sequences, the performance is significantly better, especially with increasing voxel size, as the problems of the regular TSDF increase as well.
In all other sequences the performance is very similar, which is also due to the fact that the fusion process makes the locally visible area compliant: given enough observations from the current viewpoint, all conflicts in the representation will be evened out due to the running average (unless the conflicting side has been observed a long time, resulting in a high weight).
\begin{figure}[!t]
  \centering
  \includegraphics[width=\linewidth]{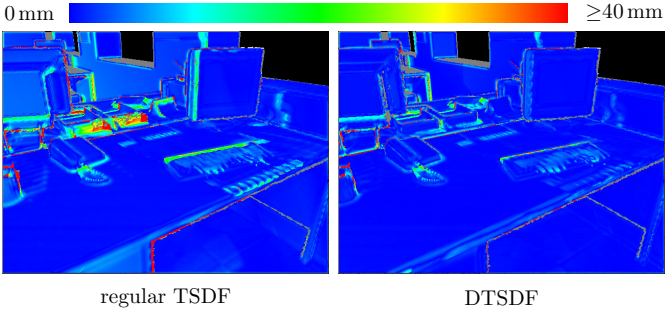} 
  \caption{Visualization of post-fusion error of frame 330 in sequence \textit{Zhou office 2}.}
  \label{fig:post_error_comparison}
\end{figure}
\begin{figure}[t]
  \centering
  \setlength{\abovecaptionskip}{1pt}
  \setlength{\belowcaptionskip}{8pt}
  \includegraphics[width=.8\linewidth]{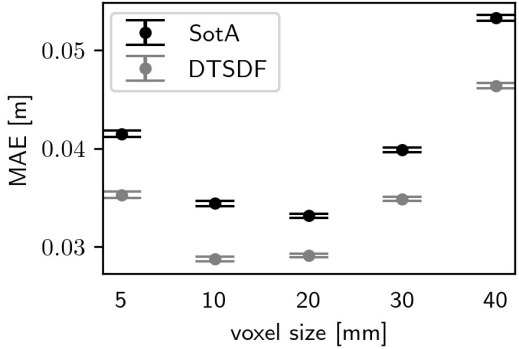}
  \caption{Post-fusion MAE (dot) and 95\% confidence intervals (bars) on example dataset \textit{TUM fr3 long office}.}
  \label{fig:icp_error}
\end{figure}
\begin{table}[!t]
  \centering
  \setlength{\belowcaptionskip}{25pt}
  \begin{tabular}{lrrrrr}
    & \multicolumn{5}{c}{voxel size [\SI{}{\mm}]} \\
    \cline{2-6}
    dataset & 5 & 10 & 20 & 30 & 40 \\
    \hline
    Zhou & 4.3 & -4.4 & -4.2 & -4.0 & -4.1 \\
    TUM\_fr1 & -2.1 & -0.7 & -0.7 & -0.7 & -0.5 \\
    TUM\_fr3 & -1.2 & -1.3 & -1.6 & -1.8 & -2.1 \\
    ICL\_NUIM & -0.2 & -2.3 & -2.1 & -2.3 & -1.4 \\
    ICL & -2.5 & -2.3 & -2.2 & -2.0 & -2.0
  \end{tabular}
  \caption{Difference of photometric MAE between SotA and DTSDF (in \%, numbers smaller zero mean DTSDF is better) for different voxel sizes averaged over all sequences in the datasets.}
  \label{tab:photometric_MAE_diff}
\end{table}
\begin{table}[!ht]
  \caption{MAE (in \SI{}{\mm}) for different voxel sizes and datasets.}
  \label{tab:mae_selection}
\resizebox{\linewidth}{!}{
  \begin{tabular}{p{70pt}p{30pt}R{24pt}R{24pt}R{24pt}R{24pt}R{24pt}}
    dataset&mode&   \multicolumn{5}{c}{voxel size [\SI{}{\mm}]} \\ \cline{3-7}
    &  & \multicolumn{1}{c}{5} & \multicolumn{1}{c}{10} & \multicolumn{1}{c}{20} & \multicolumn{1}{c}{30} & \multicolumn{1}{c}{40} \\ \hline
    \multirow{2}{*}{\shortstack[l]{SUN copyroom}}
    & SoTA & 23.0 & \textbf{20.9} & 24.8 & 54.4 & 41.8 \\
    & DTSDF & \textbf{20.7} & 30.5 & \textbf{21.7} & \textbf{25.9} & \textbf{30.0} \\
    \hline
    \multirow{2}{*}{\shortstack[l]{SUN lounge}}
    & SoTA & 16.6 & 16.9 & 24.1 & 33.2 & 45.7 \\
    & DTSDF & \textbf{14.9} & \textbf{15.7} & \textbf{21.8} & \textbf{30.0} & \textbf{39.2} \\
    \hline
    \multirow{2}{*}{\shortstack[l]{ICL NUIM\\lr kt1n}}
    & SoTA & 13.2 & 48.4 & 99.0 & 104.5 & 117.7 \\
    & DTSDF & \textbf{6.8} & \textbf{14.9} & \textbf{37.4} & \textbf{81.6} & \textbf{76.6} \\
    \hline
    \multirow{2}{*}{\shortstack[l]{ICL NUIM\\office kt3}}
    & SoTA & \textbf{2.0} & \textbf{2.2} & \textbf{3.6} & \textbf{5.5} & \textbf{7.3} \\
    & DTSDF & \textbf{2.0} & \textbf{2.2} & 3.7 & 5.6 & \textbf{7.3} \\
    \hline
    \multirow{2}{*}{\shortstack[l]{Zhou office2}}
    & SoTA & 5.5 & 9.0 & 21.9 & 52.7 & -- \\
    & DTSDF & \textbf{5.1} & \textbf{7.2} & \textbf{13.8} & \textbf{22.2} & \textbf{36.2} \\
    \hline
    \multirow{2}{*}{\shortstack[l]{turbine blade}}
    & SoTA & 2.6 & 4.9 & 13.5 & 25.0 & 34.8 \\
    & DTSDF & \textbf{2.3} & \textbf{3.1} & \textbf{5.7} & \textbf{10.5} & \textbf{17.2} \\
    \hline
    \multirow{2}{*}{\shortstack[l]{TUM fr1 desk1}}
    & SoTA & 26.7 & 23.9 & 25.9 & 31.9 & 39.3 \\
    & DTSDF & \textbf{25.2} & \textbf{22.3} & \textbf{24.6} & \textbf{29.2} & \textbf{36.0} \\
    \hline
    \multirow{2}{*}{\shortstack[l]{TUM fr3\\long office}}
    & SoTA & 41.5 & 34.4 & 33.2 & 39.8 & 53.3 \\
    & DTSDF & \textbf{35.3} & \textbf{28.8} & \textbf{29.1} & \textbf{34.9} & \textbf{46.4} \\
  \end{tabular}
}
\end{table}
In many cases this does not become apparent during tracking, but for the reusability of the overwritten parts of the completed map it is important to test these effects. To this end, we propose the geometric post-fusion per-frame error: after completing fusion of the entire sequence, for every estimated pose, a depth map is rendered again and compared to the corresponding input depth image by computing the pixel-wise mean absolute error (MAE).
 \figref{fig:post_error_comparison} shows a side-by-side example of post-fusion error images, where the regular TSDF clearly shows more errors at corners and around thin structures. \figref{fig:icp_error} plots the MAE and shows that although the tracking performance is very similar, the DTSDF is better at retaining the map.

\begin{figure}[!t]
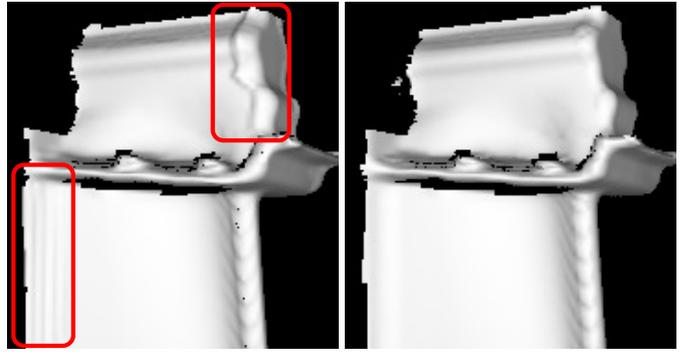

  \setlength{\belowcaptionskip}{0pt}
  \subcaptionbox{regular TSDF\label{subfig:quality_turbine_regular}}[.49\linewidth]
  {
    \centering
    \begin{tikzpicture}
      \node[anchor=south west,inner sep=0] (image) at (0,0) {\includegraphics[width=\linewidth,trim={230px 190 230 100},clip]{images/turbine_def_0680.png}};
      \begin{scope}[x={(image.south east)},y={(image.north west)}]
        \draw[red,ultra thick,rounded corners] (0.02,0.01) rectangle (0.20,0.53);
        \draw[red,ultra thick,rounded corners] (0.62,0.60) rectangle (0.85,0.99);
      \end{scope}
    \end{tikzpicture}
  }
  \subcaptionbox{DTSDF\label{subfig:quality_turbine_DTSDF}}[.49\linewidth]
  {
    \centering
    \begin{tikzpicture}
      \node[anchor=south west,inner sep=0] (image) at (0,0) {\includegraphics[width=\linewidth,trim={230px 190 230 100},clip]{images/turbine_dir_0680.png}};
    \end{tikzpicture}
  }
  \hfill
  \caption{Qualitative comparison of regular TSDF and DTSDF on turntable-style sequences. The lower left rectangle highlights artifacts from data fused from the backside. The upper right rectangle shows artifacts resulting from fusion conflicts between right- and front side.}
  \label{fig:quality_turbine}
\end{figure}

\tabref{tab:mae_selection} lists the geometric post-fusion MAE of a selection of sequences (a complete list can be found in the Appendix~\tabref{tab:mae_datasets}). One can observe that in most cases there is not too much difference, especially in concave rooms like the ICL sequences. The effect usually only affects small parts of the model, like a corner or a computer monitor. Consequently, for the mean error over the whole image the effect is not that significant, but visible nonetheless.
Object-scanning type sequence with the camera orbiting around objects generally seem to profit from the DTSDF more (c.f.~\figref{fig:quality_turbine}). Decreasing the voxel size certainly does mitigate some of these issues for the SotA, but ultimately the effect highlighted by the lower left rectangle remains for thin surfaces. Also, as \tabref{tab:tracking_ICL},~\ref{tab:tracking_tum} show, halving the voxel size instead of using the DTSDF will require more memory and computation time.

\begin{figure*}[]
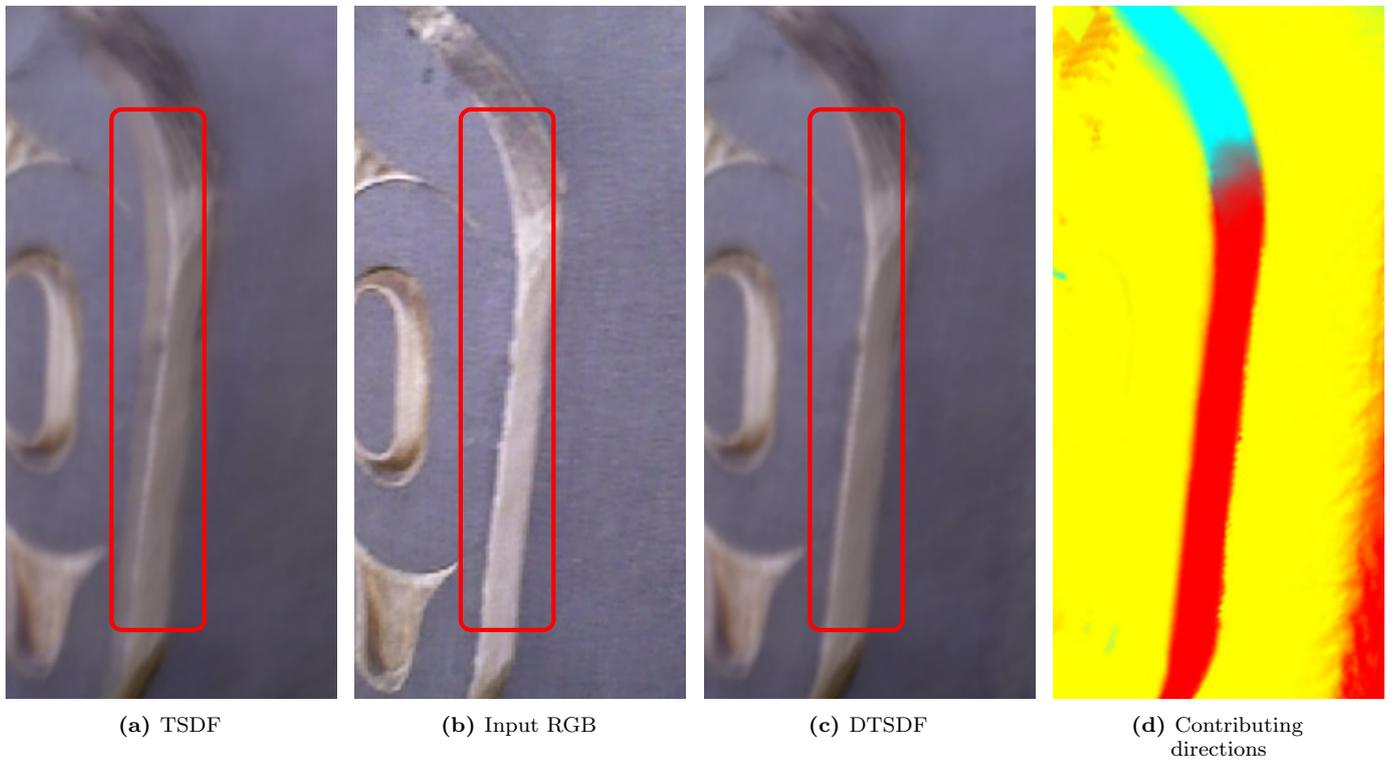

  \centering
  \subcaptionbox{TSDF\label{subfig:color_bleed_TSDF}}[.24\linewidth]
  {
    \centering
    \begin{tikzpicture}
      \node[anchor=south west,inner sep=0] (image) at (0,0) {\includegraphics[width=\linewidth,trim={230px 50 230 50},clip]{images/totempole_rgb_def_1428.png}};
      \begin{scope}[x={(image.south east)},y={(image.north west)}]
        \draw[red,ultra thick,rounded corners] (0.32,0.10) rectangle (0.60,0.85);
      \end{scope}
    \end{tikzpicture}
  }
  \hfill
  \subcaptionbox{Input RGB\label{subfig:color_bleed_rgb}}[.24\linewidth]
  {
    \centering
    \begin{tikzpicture}
      \node[anchor=south west,inner sep=0] (image) at (0,0) {\includegraphics[width=\linewidth,trim={230px 50 230 50},clip]{images/totempole_input_rgb_1428.png}};
      \begin{scope}[x={(image.south east)},y={(image.north west)}]
        \draw[red,ultra thick,rounded corners] (0.32,0.10) rectangle (0.60,0.85);
      \end{scope}
    \end{tikzpicture}
  }
  \hfill
  \subcaptionbox{DTSDF\label{subfig:color_bleed_DTSDF}}[.24\linewidth]
  {
    \centering
    \begin{tikzpicture}
      \node[anchor=south west,inner sep=0] (image) at (0,0) {\includegraphics[width=\linewidth,trim={230px 50 230 50},clip]{images/totempole_rgb_dir_1428.png}};
      \begin{scope}[x={(image.south east)},y={(image.north west)}]
        \draw[red,ultra thick,rounded corners] (0.32,0.10) rectangle (0.60,0.85);
      \end{scope}
    \end{tikzpicture}
  }
  \hfill
  \subcaptionbox{Contributing\\directions\label{subfig:color_bleed_directions}}[.24\linewidth]
  {
    \centering
    \begin{tikzpicture}
      \node[anchor=south west,inner sep=0] (image) at (0,0) {\includegraphics[width=\linewidth,trim={230px 50 230 50},clip]{images/totempole_directions_1428.png}};
    \end{tikzpicture}

  }
  \caption{Color bleeding effect on Stanford totempole sequence.}
  \label{fig:color_bleed}
\end{figure*}

Regarding the improvements of color fusion and rendering, \figref{fig:color_bleed} gives a good example the DTSDF's advantage in color separation. While in the regular TSDF (\figref{subfig:color_bleed_TSDF}) the colors blend because of fusion from two surfaces into the same voxels, the DTSDF retains different colors across edges (\figref{subfig:color_bleed_DTSDF}). \figref{subfig:color_bleed_directions} shows which directions contribute to which rendered pixel.
To quantify these improvements, we compute the photometric post-fusion per-frame MAE, analogue to the geometric error. The results, averaged over the datasets, are presented in~\tabref{tab:photometric_MAE_diff}. The effect is most visible at specific locations, but even the averaged error improves.

\subsection{Photometric ICP}
\label{ssec:evaluation_photometric}
In this subsection, we analyze the impact of photometric ICP and especially compare the three reference modes \ftof, \ftokf~and \ftor. Again, all comparisons use the RMSE RPE.

\begin{figure}[ht]
  \setlength{\abovecaptionskip}{0pt}
  \centering
  \includegraphics[width=\linewidth]{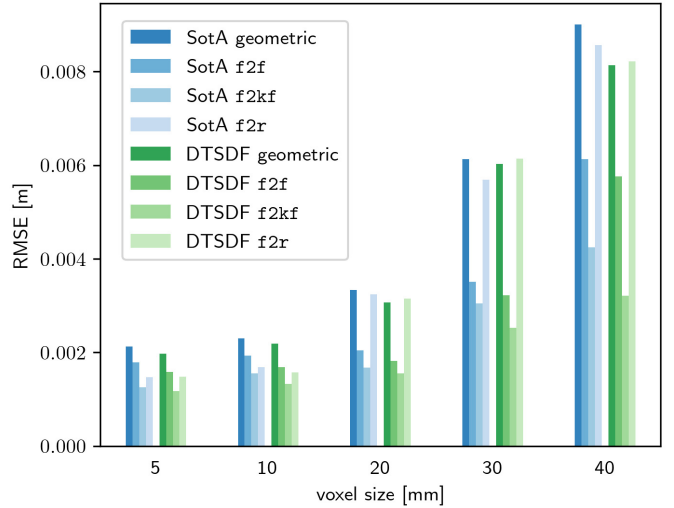}
  \caption{RPE of different tracking modes for sequence \emph{ICL diamond walk}.}
  \label{fig:RPE_color_mode}
\end{figure}

In \tabref{tab:win_loss_tsdf_mode} we analyze over all datasets, whether the DTSDF or the regular TSDF has better tracking by means of a win-loss-tie table. We use a 0.1\% tie rate, so if the RPEs differ by no more than 0.1\% it's considered a tie. The results clearly show, that the DTSDF is the overall winner and in combination with photometric tracking, the gap is even higher. In the Zhou dataset, the DTSDF outperforms the SotA in most sequences. Especially in the noisy sequences, the regular TSDF fails completely at larger voxel sizes.
Some sequences also confirm our hypothesis that the tracking performance of \ftor~degrades more with increased voxel size than the other variants. In \figref{fig:RPE_color_mode}, the gap between \ftor~and the other modes increases with growing voxel size.

\begin{table}[H]
  \centering
\resizebox{\linewidth}{!}{
  \begin{tabular}{p{5mm}lcC{10mm}C{10mm}C{10mm}}
    \shortstack{voxel\\size\\$[$\SI{}{mm}$]$} & \shortstack{TSDF\\mode} & \shortstack{mode win\\\ftof--\ftokf--\ftor}
    & \shortstack{mean\\$\frac{\mathrm{RPE}(\text{\ftof})}{\mathrm{RPE}(\text{\texttt{geom}})}$}
    & \shortstack{mean\\$\frac{\mathrm{RPE}(\text{\ftokf})}{\mathrm{RPE}(\text{\texttt{geom}})}$}
    & \shortstack{mean\\$\frac{\mathrm{RPE}(\text{\ftor})}{\mathrm{RPE}(\text{\texttt{geom}})}$} \\
    \hline
    \multirow{2}*{{5}} & SotA & \makebox[3mm][r]{9}\,--\makebox[4mm][c]{\textbf{14}}--\,\makebox[3mm][l]{10} & 1.512 & 1.412 & \textbf{1.029} \\
    & DTSDF & \makebox[3mm][r]{6}\,--\makebox[4mm][c]{\textbf{14}}--\,\makebox[3mm][l]{13} & 1.322 & 1.083 & \textbf{1.006} \\
    \hline
    \multirow{2}*{{10}} & SotA & \makebox[3mm][r]{4}\,--\makebox[4mm][c]{\textbf{16}}--\,\makebox[3mm][l]{13} & 1.053 & \textbf{0.990} & 1.015 \\
    & DTSDF & \makebox[3mm][r]{5}\,--\makebox[4mm][c]{\textbf{18}}--\,\makebox[3mm][l]{10} & 1.102 & \textbf{0.886} & 1.028 \\
    \hline
    \multirow{2}*{{20}} & SotA & \makebox[3mm][r]{6}\,--\makebox[4mm][c]{\textbf{16}}--\,\makebox[3mm][l]{11} & 0.976 & \textbf{0.894} & 1.117 \\
    & DTSDF & \makebox[3mm][r]{7}\,--\makebox[4mm][c]{\textbf{19}}--\,\makebox[3mm][l]{7} & 1.023 & \textbf{0.914} & 1.293 \\
    \hline
    \multirow{2}*{{30}} & SotA & \makebox[3mm][r]{6}\,--\makebox[4mm][c]{\textbf{17}}--\,\makebox[3mm][l]{10} & 0.939 & \textbf{0.905} & 0.965 \\
    & DTSDF & \makebox[3mm][r]{7}\,--\makebox[4mm][c]{\textbf{17}}--\,\makebox[3mm][l]{9} & 0.927 & \textbf{0.907} & 1.109 \\
    \hline
    \multirow{2}*{{40}} & SotA & \makebox[3mm][r]{5}\,--\makebox[4mm][c]{\textbf{17}}--\,\makebox[3mm][l]{9~\tablefootnote{\label{fn:missing_sequence}Tracking failed for some sequences at this resolution.}} & 1.027 & \textbf{0.989} & 1.329 \\
    & DTSDF & \makebox[3mm][r]{6}\,--\makebox[4mm][c]{\textbf{21}}--\,\makebox[3mm][l]{5~\footnoteref{fn:missing_sequence}} & 1.166 & \textbf{0.931} & 1.096
  \end{tabular}
}
  \caption{Comparison of ICP results of photometric modes \ftof, \ftokf~and \ftor~for different voxel sizes. The mode win column shows how many of the individual sequences are dominated by which mode. The rightmost three columns show the mean of the RPE ratios between \ftof, \ftokf~and \ftor~w.r.t. geometric ICP (smaller is better).}
  \label{tab:win_loss_color_mode}
\end{table}

\begin{table}[h]
  \centering
\resizebox{\linewidth}{!}{
  \begin{tabular}{lp{1.4cm}p{1.4cm}p{1.4cm}p{1.4cm}}
    dataset & \shortstack{geometric\\ICP\\\vspace{1.32mm}} & \shortstack{combined\\ICP\\\ftof} & \shortstack{combined\\ICP\\\ftokf} & \shortstack{combined\\ICP\\\ftor} \\
    \hline
Zhou & \makebox[3mm][r]{1}--\makebox[4mm][c]{\textbf{19}}--\makebox[3mm][l]{0} & \makebox[3mm][r]{1}--\makebox[4mm][c]{\textbf{19}}--\makebox[3mm][l]{0} & \makebox[3mm][r]{1}--\makebox[4mm][c]{\textbf{18}}--\makebox[3mm][l]{0}& \makebox[3mm][r]{4}--\makebox[4mm][c]{\textbf{16}}--\makebox[3mm][l]{0} \\
TUM fr3 & \makebox[3mm][r]{3}--\makebox[4mm][c]{\textbf{21}}--\makebox[3mm][l]{1} & \makebox[3mm][r]{4}--\makebox[4mm][c]{\textbf{19}}--\makebox[3mm][l]{2} & \makebox[3mm][r]{7}--\makebox[4mm][c]{\textbf{18}}--\makebox[3mm][l]{0} & \makebox[3mm][r]{8}--\makebox[4mm][c]{\textbf{17}}--\makebox[3mm][l]{0} \\
ICL NUIM & \makebox[3mm][r]{\textbf{32}}--\makebox[4mm][c]{31}--\makebox[3mm][l]{17} & \makebox[3mm][r]{35}--\makebox[4mm][c]{\textbf{36}}--\makebox[3mm][l]{9} & \makebox[3mm][r]{\textbf{34}}--\makebox[4mm][c]{\textbf{34}}--\makebox[3mm][l]{12} & \makebox[3mm][r]{\textbf{42}}--\makebox[4mm][c]{33}--\makebox[3mm][l]{5} \\
ICL & \makebox[3mm][r]{11}--\makebox[4mm][c]{\textbf{26}}--\makebox[3mm][l]{3} & \makebox[3mm][r]{12}--\makebox[4mm][c]{\textbf{26}}--\makebox[3mm][l]{2} & \makebox[3mm][r]{12}--\makebox[4mm][c]{\textbf{26}}--\makebox[3mm][l]{2} & \makebox[3mm][r]{15}--\makebox[4mm][c]{\textbf{23}}--\makebox[3mm][l]{2} \\
    \hline
total & \makebox[3mm][r]{47}--\makebox[4mm][c]{\textbf{97}}--\makebox[3mm][l]{21} & \makebox[3mm][r]{52}--\makebox[4mm][c]{\textbf{100}}--\makebox[3mm][l]{13} & \makebox[3mm][r]{54}--\makebox[4mm][c]{\textbf{96}}--\makebox[3mm][l]{15} & \makebox[3mm][r]{69}--\makebox[4mm][c]{\textbf{89}}--\makebox[3mm][l]{7}
  \end{tabular}
}
  \caption{RPE Win-loss-tie table for different datasets (format: SotA--DTSDF--tie).}
  \label{tab:win_loss_tsdf_mode}
\end{table}
It is not straightforward to answer, which of the three reference methods is the definite winner, as the performance of the overall system is rated, which depends on many factors and small differences in one module can have a huge impact on the overall trajectory. \tabref{tab:win_loss_color_mode} gives, however, a good indication: throughout all voxel sizes, the frame-to-keyframe method wins in the majority of sequences. In terms of the mean improvement over geometric ICP. Here, again, \ftokf~is the winner in most cases. Note, how \ftor~on average actually decreases tracking performance for voxel sizes larger than \SI{5}{mm}, though individual sequences do profit from it.
This shows that frame-to-keyframe offers good improvements over the other methods --- even with the simple selection scheme of every 10th frame. Frame-to-frame is a good default choice, as it is often not far behind the frame-to-keyframe tracking and, thus, no attention to keyframe selection has to be paid. The drift typically experienced with frame-to-frame tracking is not so prominent, likely because of the combined ICP, i.e., the geometric frame-to-model tracking prevents excessive drift. Frame-to-render seems to be the overall worst, especially at lower resolutions. Without proper color equalization and lighting compensation the color inside the TSDF becomes unusable for tracking, as can be observed in \figref{fig:color_mode_error_comparison}.

\subsection{Runtime and Memory Consumption}
\begin{figure}[ht]
  \setlength{\abovecaptionskip}{0pt}
  \centering
  \begin{tikzpicture}
    \node at (0, 0) {\includegraphics[width=\linewidth]{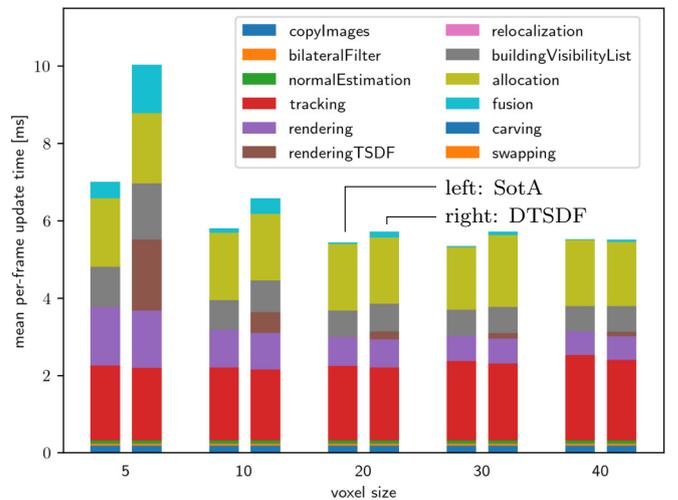}};
    \draw (0.10, 0.3) -- (0.10, 0.9) -- (1.3, 0.9) node[anchor=west, align=left] {\small{left: SotA}};
    \draw (0.65, 0.4) -- (0.65, 0.5) -- (1.3, 0.5) node[anchor=west, align=left] {\small{right: DTSDF}};
  \end{tikzpicture}
  \caption{Mean per-frame update time comparison between state-of-the-art and DTSDF on SUN totempole sequence for different voxels sizes. Conditional combination is activated.}
  \label{fig:runtime_totempole}
\end{figure}

All experiments were performed on an Intel i7-8700K with 3.70GHz and a GeForce RTX 3090. The CPU part of the code runs entirely on a single core. By reducing the changes to rendering the DTSDF while keeping the rest of the pipeline original, the runtime only differs in allocation, fusion and rendering. \figref{fig:runtime_totempole} breaks down and compares runtimes for different voxels sizes. One can observe that the additional overhead is quite small. Note that for this example conditional combination was activated, and a rendering TSDF was computed for 35-40\% of the frames with the conditions specified in~\eqref{eq:condition_1}-(\ref{eq:condition_4}).

\begin{figure}
  \centering
  \includegraphics[width=\linewidth]{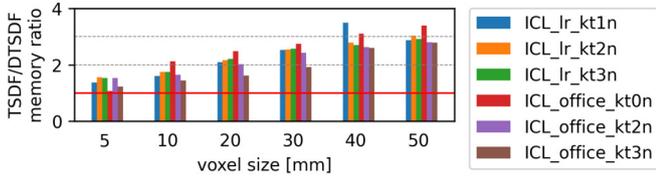}
  \caption{Ratio of allocated memory for DTSDF w.r.t. regular TSDF for different voxel sizes and scenes form the ICL NUIM dataset~\cite{Handa2014}.}
  \label{fig:allocation_ratio_ICL}
\end{figure}

As expected, the memory usage of the DTSDF is higher, as surfaces can overlap in up to three directions. In~\figref{fig:allocation_ratio_ICL} the ratio of additional memory required by the DTSDF w.r.t. the regular TSDF is displayed for ICL NUIM scenes.
For smaller voxels, the amount for extra memory is quite small. With increasing voxel size, the ratio increases as blocks are allocated in chunks of $8\times8\times8$ voxels and more surfaces with different orientations fall into the same block, though the actual number of blocks is significantly smaller.

\begin{figure}
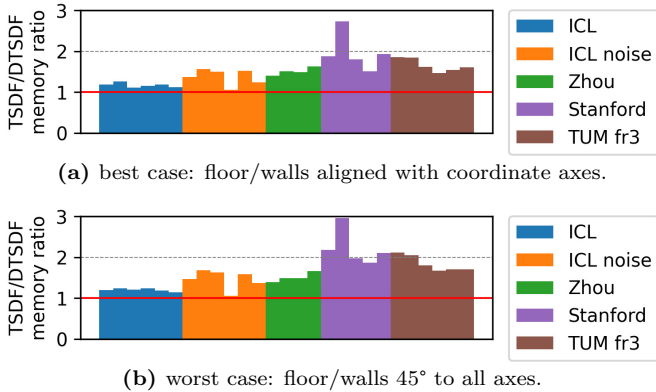

  \centering
  \begin{subfigure}[b]{\linewidth}
    \setlength{\abovecaptionskip}{-2pt}
    \setlength{\belowcaptionskip}{3pt}
    \includegraphics[width=\linewidth]{images/allocation_ratio_best_case.png}
    \caption{best case: floor/walls aligned with coordinate axes.}
    \label{fig:allocation_best_case}
  \end{subfigure}
  \begin{subfigure}[b]{\linewidth}
    \setlength{\abovecaptionskip}{-2pt}
    \setlength{\belowcaptionskip}{3pt}
    \includegraphics[width=\linewidth]{images/allocation_ratio_worst_case.png}
    \caption{worst case: floor/walls \ang{45} to all axes.}
    \label{fig:allocation_worst_case}
  \end{subfigure}
  \caption{Ratio of allocated memory for DTSDF w.r.t. regular TSDF (red line) for various sequences of different datasets and \SI{5}{mm} voxel size.}
  \label{fig:allocation_ratio_datasets}
\end{figure}

\figref{fig:allocation_ratio_datasets} plots the memory ratio for various sequences of the datasets fused with \SI{5}{mm} voxel size. It is also noticeable, that synthetic datasets (ICL, Zhou) use less memory than real ones. This is probably noise-related, as the depth-noise augmented ICL sequences also have a higher ratio, which suggests that with a more conservative allocation scheme memory can be saved. At the moment even for stray measurements blocks are allocated, as long as they have a valid normal.
As surfaces can be fused into up to three directions, determined by \eqref{eq:direction_weight}, an interesting question is how the alignment of the map coordinate frame to the scene affects memory usage.
For this, we pre-computed initial poses for each sequence by identifying the largest planes of the sequence.
In the best case scenario (\figref{fig:allocation_best_case}), the coordinate axes are parallel; in the worst case (\figref{fig:allocation_worst_case}), tilted \ang{45} to the identified planes. Noticeably, the alignment does have an effect, but not very significant. This is, however, highly dependent on the scenes and voxel sizes, because it induces only a one-time cost. Automating this process at the initialization phase of mapping is recommended.

Overall for the majority of scenes the DTSDF requires around 1.5 to 2 times as much memory as the regular TSDF.

\section{Conclusion}
\label{sec:conclusion}

In this work, we introduced the tools to use the DTSDF as a drop-in replacement for regular TSDF for mapping, tracking, and visualization applications. The ability to simply extract a regular TSDF for a given pose enables using it for a variety of tasks and with many algorithms that have been developed over the years.

We have shown that the DTSDF has advantages over the state-of-the-art for the majority of sequences, both qualitatively and quantitatively. Moreover, with the post-fusion MAE metric we showed that, while the regular TSDF is usable for local maps, reusability and revisiting of mapped places becomes problematic, if conflicting information from different surfaces corrupts the model. With reasonable memory and computation overhead, better results and a more consistent map can be obtained by the proposed DTSDF method.
Our investigation and derivation of frame-to-keyframe photometric ICP has shown that it has clear benefits over frame-to-frame and frame-to-render tracking. $\Sim3$ pose refinement shows promising first results for the crop reconstruction project.

\section*{Acknowledgments}
This work has been partially funded by the Deutsche Forschungsgemeinschaft (DFG, German Research Foundation) under Germany’s Excellence Strategy -- EXC 2070 -- 390732324 -- PhenoRob. 

\appendix
\section*{Appendix}
  \begin{table}[!hb]
    \caption{\label{tab:mae_datasets}MAE (in \SI{}{\mm}) of state-of-the-art and DTSDF compared for different voxel sizes and datasets.}
\resizebox{\linewidth}{!}{
  \begin{tabular}{p{70pt}p{30pt}R{24pt}R{24pt}R{24pt}R{24pt}R{24pt}}
    dataset&mode&   \multicolumn{5}{c}{voxel size [\SI{}{\mm}]} \\ \cline{3-7}
    &  & \multicolumn{1}{c}{5} & \multicolumn{1}{c}{10} & \multicolumn{1}{c}{20} & \multicolumn{1}{c}{30} & \multicolumn{1}{c}{40} \\ \hline
\multirow{2}*{SUN burghers}
& SoTA & \textbf{103.2} & \textbf{21.5} & 19.9 & 27.4 & 41.0 \\
& DTSDF & -- & 22.2 & \textbf{18.3} & \textbf{24.3} & \textbf{33.9} \\
\hline
\multirow{2}{*}{\shortstack[l]{SUN copyroom}}
& SoTA & 23.0 & \textbf{20.9} & 24.8 & 54.4 & 41.8 \\
& DTSDF & \textbf{20.7} & 30.5 & \textbf{21.7} & \textbf{25.9} & \textbf{30.0} \\
\hline
\multirow{2}{*}{\shortstack[l]{SUN cactusgarden}}
& SoTA & \textbf{28.8} & \textbf{27.2} & \textbf{37.5} & 52.8 & 69.9 \\
& DTSDF & 29.8 & 29.1 & 39.7 & \textbf{50.9} & \textbf{59.5} \\
\hline
\multirow{2}{*}{\shortstack[l]{SUN lounge}}
& SoTA & 16.6 & 16.9 & 24.1 & 33.2 & 45.7 \\
& DTSDF & \textbf{14.9} & \textbf{15.7} & \textbf{21.8} & \textbf{30.0} & \textbf{39.2} \\
\hline
\multirow{2}{*}{\shortstack[l]{SUN stonewall}}
& SoTA & 52.7 & \textbf{22.1} & 62.7 & \textbf{33.5} & 111.8 \\
& DTSDF & \textbf{26.1} & 60.8 & \textbf{20.7} & 66.6 & \textbf{72.6} \\
\hline
\multirow{2}{*}{\shortstack[l]{SUN totempole}}
& SoTA & \textbf{10.2} & 9.9 & 12.5 & 17.4 & 24.1 \\
& DTSDF & 10.4 & \textbf{9.5} & \textbf{10.9} & \textbf{14.5} & \textbf{19.2} \\
\hline
\multirow{2}{*}{\shortstack[l]{armadillo}}
& SoTA & 3.6 & 4.0 & 10.2 & 20.0 & 32.5 \\
& DTSDF & \textbf{3.5} & \textbf{3.6} & \textbf{6.2} & \textbf{11.1} & \textbf{17.8} \\
\hline
\multirow{2}{*}{\shortstack[l]{bunny}}
& SoTA & 2.1 & 2.5 & 8.4 & 17.7 & 27.5 \\
& DTSDF & \textbf{1.9} & \textbf{2.2} & \textbf{4.7} & \textbf{9.1} & \textbf{14.6} \\
\hline
\multirow{2}{*}{\shortstack[l]{dragon}}
& SoTA & 4.4 & 5.5 & 14.3 & 25.8 & 34.1 \\
& DTSDF & \textbf{4.1} & \textbf{4.4} & \textbf{8.8} & \textbf{16.1} & \textbf{23.0} \\
\hline
\multirow{2}{*}{\shortstack[l]{turbine blade}}
& SoTA & 2.6 & 4.9 & 13.5 & 25.0 & 34.8 \\
& DTSDF & \textbf{2.3} & \textbf{3.1} & \textbf{5.7} & \textbf{10.5} & \textbf{17.2} \\
\hline
\multirow{2}{*}{\shortstack[l]{Zhou lr1}}
& SoTA & 3.8 & 3.2 & 6.2 & 11.1 & 16.0 \\
& DTSDF & \textbf{2.7} & \textbf{3.1} & \textbf{5.7} & \textbf{10.3} & \textbf{14.9} \\
\hline
\multirow{2}{*}{\shortstack[l]{Zhou lr2}}
& SoTA & 3.8 & 4.6 & 9.0 & 14.3 & 19.8 \\
& DTSDF & \textbf{3.6} & \textbf{4.3} & \textbf{7.8} & \textbf{13.0} & \textbf{18.3} \\
\hline
\multirow{2}{*}{\shortstack[l]{Zhou office1}}
& SoTA & 4.4 & 6.5 & 18.3 & 56.4 & 58.5 \\
& DTSDF & \textbf{4.2} & \textbf{5.4} & \textbf{10.1} & \textbf{17.4} & \textbf{26.2} \\
\hline
\multirow{2}{*}{\shortstack[l]{Zhou office2}}
& SoTA & 5.5 & 9.0 & 21.9 & 52.7 & -- \\
& DTSDF & \textbf{5.1} & \textbf{7.2} & \textbf{13.8} & \textbf{22.2} & \textbf{36.2} \\
\hline
\multirow{2}{*}{\shortstack[l]{ICL NUIM\\lr kt0}}
& SoTA & \textbf{2.3} & \textbf{2.9} & 6.1 & 7.5 & \textbf{9.5} \\
& DTSDF & 2.6 & \textbf{2.9} & \textbf{5.8} & \textbf{7.4} & \textbf{9.5} \\
\hline
\multirow{2}{*}{\shortstack[l]{ICL NUIM\\lr kt1}}
& SoTA & \textbf{2.1} & \textbf{2.1} & 2.8 & 4.6 & 6.2 \\
& DTSDF & \textbf{2.1} & \textbf{2.1} & \textbf{2.7} & \textbf{4.4} & \textbf{6.0} \\
\hline
\multirow{2}{*}{\shortstack[l]{ICL NUIM\\lr kt2}}
& SoTA & \textbf{3.4} & \textbf{3.6} & \textbf{5.3} & 8.5 & 12.3 \\
& DTSDF & \textbf{3.4} & 3.7 & \textbf{5.3} & \textbf{8.3} & \textbf{12.0} \\
\hline
\multirow{2}{*}{\shortstack[l]{ICL NUIM\\lr kt3}}
& SoTA & \textbf{71.9} & \textbf{31.5} & \textbf{41.5} & 234.4 & \textbf{25.5} \\
& DTSDF & 89.8 & 83.1 & 69.0 & \textbf{152.7} & 124.0 \\
\hline
\multirow{2}{*}{\shortstack[l]{ICL NUIM\\lr kt0n}}
& SoTA & 6.6 & 7.1 & 8.5 & 12.1 & 13.0 \\
& DTSDF & \textbf{6.3} & \textbf{6.5} & \textbf{8.4} & \textbf{10.1} & \textbf{11.8} \\
\hline
\multirow{2}{*}{\shortstack[l]{ICL NUIM\\lr kt1n}}
& SoTA & 13.2 & 48.4 & 99.0 & 104.5 & 117.7 \\
& DTSDF & \textbf{6.8} & \textbf{14.9} & \textbf{37.4} & \textbf{81.6} & \textbf{76.6} \\
\hline
\multirow{2}{*}{\shortstack[l]{ICL NUIM\\lr kt2n}}
& SoTA & 22.1 & 47.4 & 87.6 & 94.9 & 95.7 \\
& DTSDF & \textbf{15.8} & \textbf{21.2} & \textbf{44.5} & \textbf{50.4} & \textbf{72.9} \\
\hline
\multirow{2}{*}{\shortstack[l]{ICL NUIM\\lr kt3n}}
& SoTA & 24.3 & 79.9 & \textbf{134.5} & \textbf{53.2} & -- \\
& DTSDF & \textbf{20.3} & \textbf{20.0} & 138.9 & 97.2 & \textbf{62.2} \\
\hline
\multirow{2}{*}{\shortstack[l]{ICL NUIM\\office kt0}}
& SoTA & \textbf{2.4} & \textbf{2.6} & 3.7 & 5.7 & \textbf{6.9} \\
& DTSDF & \textbf{2.4} & \textbf{2.6} & \textbf{3.6} & \textbf{5.6} & \textbf{6.9} \\
\hline
\multirow{2}{*}{\shortstack[l]{ICL NUIM\\office kt1}}
& SoTA & \textbf{1.1} & \textbf{1.2} & \textbf{1.8} & \textbf{2.6} & \textbf{4.0} \\
& DTSDF & \textbf{1.1} & \textbf{1.2} & \textbf{1.8} & \textbf{2.6} & 4.1 \\
\hline
\multirow{2}{*}{\shortstack[l]{ICL NUIM\\office kt2}}
& SoTA & \textbf{2.6} & 42.3 & 3.9 & 5.9 & 70.6 \\
& DTSDF & 2.7 & \textbf{2.7} & \textbf{3.7} & \textbf{5.8} & \textbf{61.6} \\
\hline
\multirow{2}{*}{\shortstack[l]{ICL NUIM\\office kt3n}}
& SoTA & 459.8 & 376.5 & 258.3 & 159.2 & \textbf{147.0} \\
& DTSDF & \textbf{83.9} & \textbf{5.8} & \textbf{6.3} & \textbf{154.3} & 159.5 \\
\hline
\end{tabular}
}
\end{table}

\captionsetup{justification=raggedright,singlelinecheck=false}

\begin{table}[!ht]
  \ContinuedFloat
  \caption{(Continued)}
\resizebox{\linewidth}{!}{
  \begin{tabular}{p{70pt}p{30pt}R{24pt}R{24pt}R{24pt}R{24pt}R{24pt}}
    dataset&mode&   \multicolumn{5}{c}{voxel size [\SI{}{\mm}]} \\ \cline{3-7}
    &  & \multicolumn{1}{c}{5} & \multicolumn{1}{c}{10} & \multicolumn{1}{c}{20} & \multicolumn{1}{c}{30} & \multicolumn{1}{c}{40} \\ \hline
\multirow{2}{*}{\shortstack[l]{TUM fr3 long\\office}}
& SoTA & 41.5 & 34.4 & 33.2 & 39.8 & 53.3 \\
& DTSDF & \textbf{35.3} & \textbf{28.8} & \textbf{29.1} & \textbf{34.9} & \textbf{46.4} \\
\hline
\multirow{2}{*}{\shortstack[l]{TUM fr3 structure\\texture far}}
& SoTA & 35.9 & \textbf{13.5} & 12.4 & 15.1 & 17.8 \\
& DTSDF & \textbf{21.9} & 13.7 & \textbf{11.5} & \textbf{12.7} & \textbf{14.1} \\
\hline
\multirow{2}{*}{\shortstack[l]{TUM fr3 structure\\texture near}}
& SoTA & 10.6 & 9.9 & 111.6 & 12.2 & 16.6 \\
& DTSDF & \textbf{8.9} & \textbf{8.6} & \textbf{9.8} & \textbf{10.2} & \textbf{13.2} \\
\hline
\multirow{2}{*}{\shortstack[l]{TUM fr3 structure\\notexture far}}
& SoTA & 15.5 & \textbf{10.7} & 11.2 & 13.5 & 18.0 \\
& DTSDF & \textbf{14.4} & 11.7 & \textbf{10.8} & \textbf{12.0} & \textbf{17.3} \\
\hline
\multirow{2}{*}{\shortstack[l]{TUM fr3 structure\\notexture near}}
& SoTA & 4.4 & 4.4 & 4.8 & 6.0 & 7.3 \\
& DTSDF & \textbf{3.8} & \textbf{3.7} & \textbf{3.9} & \textbf{4.3} & \textbf{5.0} \\
\hline
\multirow{2}{*}{\shortstack[l]{TUM fr1 desk1}}
& SoTA & 26.7 & 23.9 & 25.9 & 31.9 & 39.3 \\
& DTSDF & \textbf{25.2} & \textbf{22.3} & \textbf{24.6} & \textbf{29.2} & \textbf{36.0} \\
\hline
\end{tabular}
}
\end{table}
\bibliography{references}

\end{document}